\PassOptionsToPackage{unicode}{hyperref}
\PassOptionsToPackage{hyphens}{url}
\documentclass[12pt]{article}
\usepackage{amsmath,amsthm,amssymb,bm,graphicx,mathrsfs}
\usepackage{subcaption,float,booktabs,diagbox,mathtools,tabularx}
\usepackage[ruled,vlined]{algorithm2e}
\usepackage[round,authoryear,sort&compress]{natbib}
\setcitestyle{open={(},close={)}}
\usepackage{xcolor}
\usepackage{hyperref}
\graphicspath{{figures/}}

\newcommand{\anon}{1}
\def\spacingset#1{\renewcommand{\baselinestretch}{#1}\small\normalsize}
\newtheorem{theorem}{Theorem}

\newtheorem{corollary}{Corollary}

\newtheorem{assumption}{Assumption}

\theoremstyle{plain}
\numberwithin{equation}{section}
\usepackage{authblk}
\usepackage{comment}

\allowdisplaybreaks[4]

\begin{document}

\if1\anon
{
  \title{\bf Tensor Recovery with Heterogeneous Mode Structures}
    \author[1]{Zhixin Huang}
    \author[2]{Ziyuan Chen}
    \author[2]{Fang Yao}
    \author[3]{Ying Yang \thanks{Ying Yang is the corresponding author. Email: yangying@fudan.edu.cn}}
    \affil[1]{School of Mathematical Sciences, Fudan University, Shanghai, China}
    \affil[2]{School of Mathematical Sciences, Peking University, Beijing, China}
    \affil[3]{Center for Applied Mathematics, School of Mathematical Sciences, Fudan University, Shanghai, China}
  \date{}
  \maketitle
}
\fi

\bigskip
\begin{abstract}
Tensor recovery methods often impose a common structural assumption across modes, even when different modes may exhibit distinct forms of regularity. We propose a unified convex framework that treats each mode according to its structure. Smoothly varying modes form an indexing domain and are handled by local kernel weighting, whereas retained modes may receive low-rank, group-sparse, graph-Laplacian, or autoregressive regularization in distinct and potentially overlapping combinations. The framework also covers ordinary tensor recovery without local smoothing. A consensus HM-ADMM algorithm separates the active penalties into tractable proximal and linear-system updates. For the population-quadratic estimator, we derive deterministic oracle inequalities and high-probability bounds under independent trace regression. The bounds disentangle stochastic error, rank and support complexity, smoothing bias, and quadratic mismatch. They also identify when quadratic penalties sharpen the Frobenius-error bound. For regular smooth domains, the theory characterizes the bias-variance trade-off and yields an optimal bandwidth order, with a specialization to uniform tensor completion. Simulations and two applications illustrate the gains from matching regularization to mode-specific structure, especially under severe undersampling.
\end{abstract}

\vfill
\newpage
\spacingset{1.8}
\setlength{\abovedisplayskip}{2.5pt}
\setlength{\belowdisplayskip}{2.5pt}
\setlength{\abovedisplayshortskip}{2.5pt}
\setlength{\belowdisplayshortskip}{2.5pt}

\section{Introduction}
\label{sec:introduction}

Tensor-valued data arise when observations are indexed by several scientifically distinct axes, such as space, time, wavelength, experimental condition or subject characteristics. Retaining these axes in tensor form preserves interactions that can be obscured by vectorization or matricization and has motivated a large literature on tensor decomposition and recovery \citep{kolda2009tensor,sidiropoulos2017tensor,yuan2016tensor}.

In many applications, only a small fraction of tensor entries is observed and the measurements may be noisy. Recovering the full tensor often requires specific assumptions on its underlying structure.
Low-rank structure has been a commonly used condition in tensor recovery. Convex approaches extended matrix nuclear norm methods to tensor unfoldings, yielding computationally tractable procedures for completion from partial observations \citep{gandy2011tensor,liu2013tensor}. Subsequent work developed tensor nuclear norms that more directly exploit multilinear low-rank structure \citep{tomioka2013structured,yuan2016tensor}. General convex regularization theory \citep{raskutti2019tensor} also clarified how low-rank structure affects estimation in high-dimensional tensor regression. These developments provide a strong foundation when low rank is the underlying tensor structure.

Beyond low rank, application-specific methods have incorporated additional forms of regularity, such as temporal smoothness in video completion \citep{wang2014spatiotemporal}, autoregressive dependence in spatio-temporal tensor factorization \citep{takeuchi2017autoregressive}, joint sparsity in tensor regression \citep{sun2017store}, and graph- or manifold-based regularization in hyperspectral tensor completion \citep{tian2026manifold}. Although these methods demonstrate the value of structure beyond low rank, they are generally developed for a particular structural configuration, combining low rank with one additional form of regularity. This limitation becomes consequential when tensor modes encode scientifically distinct sources of variation. A single tensor may vary smoothly along one mode, have low-rank structure along another, be sparse along a third, and exhibit graph or autoregressive structure along others. We use the term \emph{tensor recovery with heterogeneous mode structures} for recovery problems in which the form of regularity exploited differs across tensor modes. In such problems, the penalties act across multiple unfoldings of the same tensor and are coupled through a common estimation error. Their analysis must reconcile decomposable degree-one penalties with positive-semidefinite quadratic penalties. Their joint treatment in optimization and recovery analysis remains underdeveloped.

To address this problem, we develop a unified convex framework for tensor recovery with heterogeneous mode structures. The treatment of each mode reflects the regularity being exploited. Modes along which the tensor varies smoothly form an indexing domain and are handled through local kernel weighting, whereas the remaining modes form the tensor-valued target, with low-rank, group-sparse, geometry-aware, and autoregressive structures encoded through mode-specific penalties in distinct and potentially overlapping combinations. These four components are representative rather than exhaustive: the framework can accommodate other decomposable degree-one and positive-semidefinite quadratic regularizers, with the corresponding proximal updates and compatibility quantities adapted accordingly. When multiple index locations are available, local weighting pools observations over a potentially multidimensional domain. The singleton case recovers ordinary tensor estimation without local borrowing. We formulate the resulting estimator through a population-quadratic criterion that preserves the population curvature while concentrating sampling variation in the empirical linear term.

Our contributions are threefold. First, we formulate tensor recovery with heterogeneous mode structures, treating smooth modes through local weighting and encoding four representative retained-mode structures, low-rank (LR), group-sparse (SP), geometry-aware (GEO), and autoregressive (AR) structures, on the retained modes through mode-specific penalties. This yields a common population-quadratic convex estimator for both smooth tensor fields and ordinary tensor recovery. Rather than defining a separate procedure for each structural configuration, the same estimator accommodates different and potentially overlapping assignments of different structures.
Second, we develop HM-ADMM as a modular computational realization of this formulation. Activating or removing a structural component changes only its corresponding auxiliary block, while leaving the remaining updates unchanged. The LR and SP blocks reduce to proximal mappings, whereas the GEO and AR blocks reduce to reusable linear-system solves.
Third, we establish deterministic oracle inequalities and high-probability recovery bounds that separate stochastic score fluctuation, rank and support complexity, smoothing bias, and mismatch with the imposed quadratic structures. They also identify conditions under which the additional curvature from GEO and AR regularization yields a strictly sharper rank- and support-dependent Frobenius-error bound than that for the corresponding estimator without quadratic penalties. For regular smooth domains, the bounds characterize the bias--variance trade-off induced by local borrowing and yield an optimal bandwidth order. 
Simulations and two real-data applications show that matching regularization to mode-specific structure can improve recovery, while local borrowing is particularly beneficial under severe undersampling.

The remainder of the paper is organized as follows.  Section~\ref{sec:model} introduces the heterogeneous-mode representation, the tensor trace-regression model, and the population-quadratic estimator.  Section~\ref{sec:algorithm} develops the HM-ADMM algorithm.  Section~\ref{sec:theory} establishes deterministic and probabilistic recovery guarantees, including smooth-domain and tensor-completion specializations.  Sections~\ref{sec:simulation} and~\ref{sec:realdata} present the simulation studies and real-data applications. 

\section{Heterogeneous-Mode Model and Estimation}\label{sec:model}

Let $\mathcal T^*\in\mathbb R^{n_1\times\cdots\times n_D}$ be an unknown order-$D$ tensor. Modes along which $\mathcal T^*$ is modeled as varying smoothly are represented as indexing coordinates, while the remaining modes form the tensor-valued target and are regularized directly.

For notational convenience, label the $k$ retained modes as the first $k$ modes. The remaining $d=D-k$ modes define the indexing coordinate $s=(s_1,\ldots,s_d)\in\mathcal S\subset\mathbb R^d$. Fixing $s$ yields \( \mathcal M^*(s)\in\mathbb R^{n_1\times\cdots\times n_k},\ s\in\mathcal S, \) and the collection $\{\mathcal M^*(s):s\in\mathcal S\}$ forms the tensor field to be recovered. The indexing domain may be one- or multidimensional. The singleton case $\mathcal S=\{s_0\}$ corresponds to ordinary tensor recovery, with $\mathcal M^*(s_0)=\mathcal T^*$. At each $s\in\mathcal S$, suppose that we observe $\{(y_{si},\mathcal X_{si}):i=1,\ldots,m_s\}$ from the tensor trace-regression model
\[
y_{si}
=
\langle\mathcal X_{si},\mathcal M^*(s)\rangle
+
\varepsilon_{si},
\qquad i=1,\ldots,m_s,
\]
where $\mathcal X_{si}\in\mathbb R^{n_1\times\cdots\times n_k}$, \( \langle\mathcal A,\mathcal B\rangle = \sum_{j_1=1}^{n_1}\cdots\sum_{j_k=1}^{n_k} \mathcal A_{j_1,\ldots,j_k} \mathcal B_{j_1,\ldots,j_k} \) denotes the Frobenius inner product between two tensors of the same dimensions and $\mathbb E(\varepsilon_{si}\mid\mathcal X_{si})=0$. Tensor completion is obtained when each $\mathcal X_{si}$ is a canonical basis tensor.

Write $|\mathcal S|$ for the cardinality of $\mathcal S$. When $|\mathcal S|>1$ and $\mathcal M^*(\cdot)$ varies smoothly over the indexing domain, observations from nearby locations can be used to improve recovery at a target location $s$. Following the local-weighting construction of \citet{chen2024dynamic}, define
\[
w_h(s',s)
=
\frac{K\{(s'-s)/h\}}
{\sum_{v\in\mathcal S}K\{(v-s)/h\}},
\qquad s'\in\mathcal S,
\]
where $K:\mathbb R^d\to\mathbb R_+$ is a nonnegative kernel and $h>0$ is a bandwidth. This construction extends local borrowing from a one-dimensional sequence of matrices to a tensor field over a potentially multidimensional domain. For $\mathcal S=\{s_0\}$, we set $w_h(s_0,s_0)=1$, so no local borrowing is performed and the bandwidth plays no role. Otherwise, the bandwidth determines the extent of local pooling and the resulting bias--variance trade-off. Thus, the same formulation exploits smooth variation through local borrowing when such structure is available and reduces to ordinary tensor recovery when no smooth structure is imposed.

\subsection{Heterogeneous structural regularization}

Fix a target location $s\in\mathcal S$. The regularizers below act on a generic candidate tensor $\mathcal M\in\mathbb R^{n_1\times\cdots\times n_k}$ for $\mathcal M^*(s)$. Because the regularization is defined pointwise in $s$, we suppress the dependence on the target location. Let $\mathcal M_{(m)}$ denote the mode-$m$ unfolding of a retained tensor $\mathcal M$. We encode structure in the retained modes through mode-specific penalties. Different modes may be assigned to different structural families, and empty structural families are allowed.  We consider four representative components, namely low-rank, mode-wise sparsity, known geometry, and autoregressive dependence, which cover the structural configurations studied in this paper.

\textit{1. Mode-wise low-rank structure.} Let $\mathcal I_{\mathrm{LR}}$ denote the set of retained modes for which a low-dimensional representation is appropriate.  We use the overlapped nuclear-norm penalty
\[
R_{\mathrm{LR}}(\mathcal M)
=
\lambda_{\mathrm{LR}}
\sum_{l\in\mathcal I_{\mathrm{LR}}}
\|\mathcal M_{(l)}\|_*,
\]
where $\|\cdot\|_*$ denotes the matrix nuclear norm.  %

\textit{2. Mode-wise group sparsity.} Let $\mathcal I_{\mathrm{SP}}$ contain the retained modes along which only a subset of indices is expected to be active.  We impose
\[
R_{\mathrm{SP}}(\mathcal M)
=
\lambda_{\mathrm{SP}}
\sum_{p\in\mathcal I_{\mathrm{SP}}}
\|\mathcal M_{(p)}\|_{2,1},
\qquad
\|\mathcal M_{(p)}\|_{2,1}
=
\sum_{j=1}^{n_p}
\|[\mathcal M_{(p)}]_{j,:}\|_2,
\]
where $\|\cdot\|_2$ denotes the Euclidean norm of a vector and $\|\cdot\|_{2,1}$ is the corresponding rowwise mixed norm. This penalty promotes sparsity across rows of the mode-$p$ unfolding, or equivalently across mode-$p$ slices of the tensor.

\textit{3. Geometry-aware structure.} Let $\mathcal I_{\mathrm{GEO}}$ index retained modes equipped with known geometric information, such as a graph, ordering, or neighborhood structure. For each $g\in\mathcal I_{\mathrm{GEO}}$, let $L_g\in\mathbb R^{n_g\times n_g}$ be a positive semidefinite operator encoding this geometry; the principal example considered here is a graph Laplacian. We use
\[
R_{\mathrm{GEO}}(\mathcal M)
=
\lambda_{\mathrm{GEO}}
\sum_{g\in\mathcal I_{\mathrm{GEO}}}
\operatorname{tr}\!\left(
\mathcal M_{(g)}^\top L_g\mathcal M_{(g)}
\right).
\]
Equivalently, defining \( \mathcal E_g^{\mathrm{GEO}}(\mathcal M) = L_g^{1/2}\mathcal M_{(g)} \) gives
\[
R_{\mathrm{GEO}}(\mathcal M)
=
\lambda_{\mathrm{GEO}}
\sum_{g\in\mathcal I_{\mathrm{GEO}}}
\|\mathcal E_g^{\mathrm{GEO}}(\mathcal M)\|_F^2,
\]
where $L_g^{1/2}$ denotes the symmetric positive semidefinite square root of $L_g$ and $\|\cdot\|_F$ is the Frobenius norm. When $L_g$ is a graph Laplacian, the penalty is the usual graph Dirichlet energy and encourages adjacent mode-$g$ slices to be similar; see \citet{dong2016laplacian,perraudin2017stationary} for related graph-spectral interpretations.

\textit{4. Autoregressive structure.} Let $a$ denote a retained temporal mode of length $T$, and let $\mathcal M_\tau$ denote the slice at time $\tau$. 
For concreteness, we consider an AR(1) structure with coefficient $|\phi|<1$, treating $\phi$ as fixed in the estimator and recovery theory. In practice, $\phi$ may be specified externally or estimated before tensor optimization. In the numerical experiments, we use the pooled adjacent-pair estimator described in Section \textcolor{black}{S.6.3} of the Supplementary Material and hold the resulting estimate fixed during each optimization run. 
Let $D_\phi\in\mathbb R^{(T-1)\times T}$ denote the corresponding difference operator.
We use
\[
R_{\mathrm{AR}}(\mathcal M)
=
\lambda_{\mathrm{AR}}
\sum_{\tau=2}^{T}
\|\mathcal M_\tau-\phi\mathcal M_{\tau-1}\|_F^2
=
\lambda_{\mathrm{AR}}
\|D_\phi\mathcal M_{(a)}\|_F^2.
\]
This penalty measures departure from AR(1) evolution along the retained temporal mode. More generally, a fixed-order AR($q$) structure can be accommodated by replacing $D_\phi$ with the banded operator associated with $\mathcal M_\tau-\sum_{j=1}^q\phi_j\mathcal M_{\tau-j}$.

For later use, group the penalties as
\[
R(\mathcal M)
=
R_{\mathrm{LR}}(\mathcal M)+R_{\mathrm{SP}}(\mathcal M),
\qquad
Q(\mathcal M)
=
R_{\mathrm{GEO}}(\mathcal M)+R_{\mathrm{AR}}(\mathcal M).
\]
The LR and SP penalties are positively homogeneous and decomposable, whereas the GEO and AR penalties are quadratic. This distinction is reflected in the recovery theory: $R$ contributes rank and support complexity, while $Q$ measures departure from the imposed geometric and autoregressive structures.

\vspace{-0.1in}
\subsection{Population-quadratic estimator}

For a fixed location $s\in\mathcal S$, a natural criterion for estimating $\mathcal M^*(s)$ is the locally weighted squared loss
\[
\widehat L_s^{\mathrm{loc}}(\mathcal M)
=
\sum_{s'\in\mathcal S}
\frac{w_h(s',s)}{m_{s'}}
\sum_{i=1}^{m_{s'}}
\left\{
y_{s'i}
-
\langle\mathcal X_{s'i},\mathcal M\rangle
\right\}^2.
\]
This criterion fits a common tensor $\mathcal M$ to observations collected near $s$, with the contribution of location $s'$ determined by $w_h(s',s)$. For the singleton domain $\mathcal S=\{s_0\}$, it reduces to the ordinary average squared loss based on the $m_{s_0}$ observations at $s_0$. To simplify subsequent expressions, let \( \mathcal J_s = \left\{ (s',i): s'\in\mathcal S,\; i=1,\ldots,m_{s'},\; w_h(s',s)>0 \right\}. \) For $j=(s',i)\in\mathcal J_s$, write \( y_j=y_{s'i},\ \mathcal X_j=\mathcal X_{s'i}, \) and \( a_{s,j} = {w_h(s',s)}/{m_{s'}}\). Thus $a_{s,j}$ is the weight assigned to an individual observation, and \( \sum_{j\in\mathcal J_s}a_{s,j}=1\). In this notation,
\[
\widehat L_s^{\mathrm{loc}}(\mathcal M)
=
\sum_{j\in\mathcal J_s}
a_{s,j}
\left\{
y_j-\langle\mathcal X_j,\mathcal M\rangle
\right\}^2.
\]

Define the empirical weighted first moment $\widehat{\mathcal B}_s=\sum_{j\in\mathcal J_s}a_{s,j}y_j\mathcal X_j$ and the population second-moment operator $\mathscr S_s(\mathcal H)=\mathbb E\!\left[\sum_{j\in\mathcal J_s}a_{s,j}\langle\mathcal X_j,\mathcal H\rangle\mathcal X_j\right]$, where the expectation is taken with respect to the design tensors. Throughout the paper, $\mathscr S_s$ is assumed to be known from, or specified by, the design distribution. Following the population-quadratic construction of \citet{chen2024dynamic}, we retain the empirical first moment $\widehat{\mathcal B}_s$ but replace the empirical quadratic operator with its population counterpart. The resulting criterion is
\[
\widetilde L_s(\mathcal M)=\langle\mathcal M,\mathscr S_s(\mathcal M)\rangle-2\langle\widehat{\mathcal B}_s,\mathcal M\rangle. 
\]
Its linear component retains the empirical weighted information in the observations, whereas its quadratic component is governed by the population design operator.  

Combining this criterion with the heterogeneous structural penalties gives
\begin{equation}
\widetilde{\mathcal M}(s)
\in
\arg\min_{\mathcal M}
\left\{
\widetilde L_s(\mathcal M)
+
R(\mathcal M)
+
Q(\mathcal M)
\right\}.
\label{eq:heterogeneous_estimator}
\end{equation}
The estimator covers both locally smoothed and ordinary tensor recovery. When $\mathcal S=\{s_0\}$, the observation weights reduce to $a_{s_0,(s_0,i)}=1/m_{s_0}$, no cross-location borrowing is performed, and we write $\widetilde{\mathcal M}=\widetilde{\mathcal M}(s_0)$. When $|\mathcal S|>1$, the weights $a_{s,j}$ determine how observations from neighboring locations contribute to recovery at $s$. In either regime, the active retained-mode structures enter through $R$ and $Q$, so their form can be varied without changing the definition of the estimator.

For fixed structural operators and autoregressive parameter $\phi$, the objective in \eqref{eq:heterogeneous_estimator} is convex whenever $\mathscr S_s$ and the $L_g$ are positive semidefinite. Under the population-curvature condition imposed in Section~\ref{sec:theory}, the objective is strongly convex and coercive, so \eqref{eq:heterogeneous_estimator} has a unique minimizer. The different structural components nevertheless act on different unfoldings of the common tensor variable, so direct optimization couples their respective proximal or quadratic updates.  Section~\ref{sec:algorithm} develops HM-ADMM to separate these components into tractable subproblems while preserving the unified estimator in \eqref{eq:heterogeneous_estimator}.

\vspace{-0.15in}
\section{HM-ADMM Algorithm}
\label{sec:algorithm}

The heterogeneous structural components in \eqref{eq:heterogeneous_estimator} act on different unfoldings of the same tensor and therefore cannot in general be optimized independently in the original formulation.  We develop a consensus alternating direction method of multipliers, termed HM-ADMM, that introduces separate auxiliary copies for the active structural components.  Conditional on the common primal tensor, these auxiliary blocks decouple into singular-value thresholding, group soft-thresholding, and quadratic linear-system updates.

The regularization parameters $\lambda_{\mathrm{LR}}$, $\lambda_{\mathrm{SP}}$, $\lambda_{\mathrm{GEO}}$, and $\lambda_{\mathrm{AR}}$ are statistical tuning parameters and remain fixed during optimization.  The augmented-Lagrangian parameter $\beta>0$ is instead algorithmic and is common to the consensus constraints. The derivation below allows all four structural families considered in this paper to be active. In a particular application, inactive components and their associated auxiliary and dual variables are simply omitted. Thus, when only two or three structural components are used, HM-ADMM reduces to the corresponding subset of updates.

\vspace{-0.1in}
\subsection{Consensus formulation}

The LR, SP, and GEO penalties may each be assigned to multiple retained modes, indexed by $\mathcal I_{\mathrm{LR}}$, $\mathcal I_{\mathrm{SP}}$, and $\mathcal I_{\mathrm{GEO}}$. We allow at most one retained temporal mode, denoted by $a$, to carry AR regularization, and let $\iota_{\mathrm{AR}}\in\{0,1\}$ indicate whether this component is active. Introduce auxiliary tensors $\mathcal Y_l$, $\mathcal Z_p$, and $\mathcal G_g$ for $l\in\mathcal I_{\mathrm{LR}}$, $p\in\mathcal I_{\mathrm{SP}}$, and $g\in\mathcal I_{\mathrm{GEO}}$, respectively, together with $\mathcal A$ when $\iota_{\mathrm{AR}}=1$. In $\mathcal Y_{l,(l)}$, the first subscript identifies the auxiliary copy associated with mode $l$, whereas the parenthesized subscript denotes the mode-$l$ unfolding of that copy. The same convention applies to $\mathcal Z_{p,(p)}$ and $\mathcal G_{g,(g)}$; $\mathcal A_{(a)}$ is the unfolding along the retained temporal mode.

The estimator in \eqref{eq:heterogeneous_estimator} is equivalently written as
{\small\begin{equation}
\begin{aligned}
\min_{\mathcal M,\{\mathcal Y_l\},\{\mathcal Z_p\},
\{\mathcal G_g\},\mathcal A}
\quad
&\widetilde L_s(\mathcal M)
+
\lambda_{\mathrm{LR}}
\sum_{l\in\mathcal I_{\mathrm{LR}}}
\|\mathcal Y_{l,(l)}\|_*
+
\lambda_{\mathrm{SP}}
\sum_{p\in\mathcal I_{\mathrm{SP}}}
\|\mathcal Z_{p,(p)}\|_{2,1}
\\
&\quad+
\lambda_{\mathrm{GEO}}
\sum_{g\in\mathcal I_{\mathrm{GEO}}}
\operatorname{tr}\!\left(
\mathcal G_{g,(g)}^\top L_g\mathcal G_{g,(g)}
\right)
+
\iota_{\mathrm{AR}}\lambda_{\mathrm{AR}}
\|D_\phi\mathcal A_{(a)}\|_F^2
\\
\text{subject to}\quad
&\mathcal Y_l=\mathcal M,
\quad l\in\mathcal I_{\mathrm{LR}},
&&\hspace{-4cm}
\mathcal Z_p=\mathcal M,
\quad p\in\mathcal I_{\mathrm{SP}},
\\
&\mathcal G_g=\mathcal M,
\quad g\in\mathcal I_{\mathrm{GEO}},
&&\hspace{-4cm}
\mathcal A=\mathcal M,
\quad \text{when }\iota_{\mathrm{AR}}=1.
\end{aligned}
\label{eq:consensus_problem}
\end{equation}}
Although several auxiliary variables appear, \eqref{eq:consensus_problem} is a two-block problem: one block is the primal tensor $\mathcal M$, and the other collects all active auxiliary copies. This grouping retains the standard two-block ADMM structure while allowing the auxiliary updates to separate conditionally on $\mathcal M$.

Let $\mathcal U_l$, $\mathcal V_p$, $\mathcal P_g$, and $\mathcal D_{\mathrm{AR}}$ be the scaled dual variables for the LR, SP, GEO, and AR constraints. The scaled augmented Lagrangian is
{\small\begin{align*}
L_\beta
={}&
\widetilde L_s(\mathcal M)
+
\lambda_{\mathrm{LR}}
\sum_{l\in\mathcal I_{\mathrm{LR}}}
\|\mathcal Y_{l,(l)}\|_*
+
\frac{\beta}{2}
\sum_{l\in\mathcal I_{\mathrm{LR}}}
\left(
\|\mathcal M-\mathcal Y_l+\mathcal U_l\|_F^2
-\|\mathcal U_l\|_F^2
\right)
\notag\\
&+
\lambda_{\mathrm{SP}}
\sum_{p\in\mathcal I_{\mathrm{SP}}}
\|\mathcal Z_{p,(p)}\|_{2,1}
+
\frac{\beta}{2}
\sum_{p\in\mathcal I_{\mathrm{SP}}}
\left(
\|\mathcal M-\mathcal Z_p+\mathcal V_p\|_F^2
-\|\mathcal V_p\|_F^2
\right)
\notag\\
&+
\lambda_{\mathrm{GEO}}
\sum_{g\in\mathcal I_{\mathrm{GEO}}}
\operatorname{tr}\!\left(
\mathcal G_{g,(g)}^\top L_g\mathcal G_{g,(g)}
\right)
+
\frac{\beta}{2}
\sum_{g\in\mathcal I_{\mathrm{GEO}}}
\left(
\|\mathcal M-\mathcal G_g+\mathcal P_g\|_F^2
-\|\mathcal P_g\|_F^2
\right)
\notag\\
&+
\iota_{\mathrm{AR}}
\left\{
\lambda_{\mathrm{AR}}\|D_\phi\mathcal A_{(a)}\|_F^2
+
\frac{\beta}{2}
\left(
\|\mathcal M-\mathcal A+\mathcal D_{\mathrm{AR}}\|_F^2
-\|\mathcal D_{\mathrm{AR}}\|_F^2
\right)
\right\}.
\end{align*}}
The scaled form displays the center of each auxiliary update and isolates the common primal tensor through which the structural blocks interact.

\subsection{Primal and structural updates}

Let $q=|\mathcal I_{\mathrm{LR}}|+|\mathcal I_{\mathrm{SP}}|+|\mathcal I_{\mathrm{GEO}}|+\iota_{\mathrm{AR}}$ denote the number of active consensus blocks. When $q=0$, no splitting is needed and $\widetilde L_s$ can be minimized directly; hence we consider $q\geq1$ below.
At iteration $r$, the $\mathcal M$ block minimizes
{\small\begin{equation}
\begin{aligned}
\mathcal M^{(r+1)}
\in\arg\min_{\mathcal M}
\Bigg\{
&\widetilde L_s(\mathcal M)
+
\frac{\beta}{2}
\sum_{l\in\mathcal I_{\mathrm{LR}}}
\|\mathcal M-\mathcal Y_l^{(r)}+\mathcal U_l^{(r)}\|_F^2
+
\frac{\beta}{2}
\sum_{p\in\mathcal I_{\mathrm{SP}}}
\|\mathcal M-\mathcal Z_p^{(r)}+\mathcal V_p^{(r)}\|_F^2
\\
&+
\frac{\beta}{2}
\sum_{g\in\mathcal I_{\mathrm{GEO}}}
\|\mathcal M-\mathcal G_g^{(r)}+\mathcal P_g^{(r)}\|_F^2
+
\frac{\iota_{\mathrm{AR}}\beta}{2}
\|\mathcal M-\mathcal A^{(r)}+\mathcal D_{\mathrm{AR}}^{(r)}\|_F^2
\Bigg\}.
\end{aligned}
\label{eq:M_subproblem}
\end{equation}}
Thus the structural regularizers enter the primal subproblem only through the current auxiliary and dual variables.

Let $N=n_1\cdots n_k$, let $\overline H_s$ be the matrix representation of $\mathscr S_s$, and write $b_s=\operatorname{vec}(\widehat{\mathcal B}_s)$. Thus \( \overline H_s = \mathbb E\!\left[ \sum_{j\in\mathcal J_s}a_{s,j} \operatorname{vec}(\mathcal X_j) \operatorname{vec}(\mathcal X_j)^\top \right] \). Define
\[
\mathcal C^{(r)}
=
\sum_{l\in\mathcal I_{\mathrm{LR}}}
(\mathcal Y_l^{(r)}-\mathcal U_l^{(r)})
+
\sum_{p\in\mathcal I_{\mathrm{SP}}}
(\mathcal Z_p^{(r)}-\mathcal V_p^{(r)})
+
\sum_{g\in\mathcal I_{\mathrm{GEO}}}
(\mathcal G_g^{(r)}-\mathcal P_g^{(r)})
+
\iota_{\mathrm{AR}}
(\mathcal A^{(r)}-\mathcal D_{\mathrm{AR}}^{(r)}).
\]
The first-order condition for \eqref{eq:M_subproblem} gives
\begin{equation}
\left(
\overline H_s+\frac{\beta q}{2}I_N
\right)
\operatorname{vec}(\mathcal M^{(r+1)})
=
b_s+\frac{\beta}{2}\operatorname{vec}(\mathcal C^{(r)}).
\label{eq:M_linear_system}
\end{equation}
Because $\overline H_s\succeq0$, the coefficient matrix is positive definite for $\beta>0$ and $q\geq1$, so the primal update is uniquely defined. For general trace regression, conjugate gradients can be applied using only the action of $\mathscr S_s$, without forming the $N\times N$ matrix explicitly.

The low-rank and sparse auxiliary variables admit closed-form proximal updates. For $l\in\mathcal I_{\mathrm{LR}}$, take a singular value decomposition $(\mathcal M^{(r+1)}+\mathcal U_l^{(r)})_{(l)}=U_l\Sigma_lV_l^\top$. Singular-value thresholding gives
\begin{equation}
\mathcal Y_{l,(l)}^{(r+1)}
=
U_l\left(
\Sigma_l-\frac{\lambda_{\mathrm{LR}}}{\beta}I
\right)_+V_l^\top,
\qquad l\in\mathcal I_{\mathrm{LR}},
\label{eq:Y_update}
\end{equation}
after which the result is folded back to tensor form. For $p\in\mathcal I_{\mathrm{SP}}$, set $\mathcal Q_p^{(r)}=\mathcal M^{(r+1)}+\mathcal V_p^{(r)}$. The group-sparse update acts rowwise on the mode-$p$ unfolding:
\begin{equation*}
[\mathcal Z_{p,(p)}^{(r+1)}]_{j,:}
=
\left(
1-
\frac{\lambda_{\mathrm{SP}}}
{\beta\|[\mathcal Q_{p,(p)}^{(r)}]_{j,:}\|_2}
\right)_+
[\mathcal Q_{p,(p)}^{(r)}]_{j,:},
\qquad p\in\mathcal I_{\mathrm{SP}},
\end{equation*}
with zero-norm rows set to zero. %

The geometry-aware and autoregressive auxiliary updates reduce to linear-system solves. For $g\in\mathcal I_{\mathrm{GEO}}$, set $\mathcal Q_g^{(r)}=\mathcal M^{(r+1)}+\mathcal P_g^{(r)}$. The quadratic GEO subproblem gives
\begin{equation*}
\mathcal G_{g,(g)}^{(r+1)}
=
\beta
\left(
\beta I_{n_g}+2\lambda_{\mathrm{GEO}}L_g
\right)^{-1}
\mathcal Q_{g,(g)}^{(r)},
\qquad g\in\mathcal I_{\mathrm{GEO}}.
\end{equation*}
When $\iota_{\mathrm{AR}}=1$, set $\mathcal Q_{\mathrm{AR}}^{(r)}=\mathcal M^{(r+1)}+\mathcal D_{\mathrm{AR}}^{(r)}$. The temporal update is
\begin{equation}
\mathcal A_{(a)}^{(r+1)}
=
\beta
\left(
\beta I_T+2\lambda_{\mathrm{AR}}D_\phi^\top D_\phi
\right)^{-1}
\mathcal Q_{\mathrm{AR},(a)}^{(r)}.
\label{eq:A_update}
\end{equation}
Both updates are implemented as linear-system solves rather than explicit matrix inversions. For fixed $\beta$, the coefficient matrices remain unchanged across iterations and can be factorized once; for AR(1), $D_\phi^\top D_\phi$ is tridiagonal.

\subsection{Stopping and implementation}

After the primal and auxiliary updates, the scaled dual variables are updated by
\begin{equation}
\begin{aligned}
\mathcal U_l^{(r+1)}
&=\mathcal U_l^{(r)}+\mathcal M^{(r+1)}-\mathcal Y_l^{(r+1)},
&&l\in\mathcal I_{\mathrm{LR}},\\
\mathcal V_p^{(r+1)}
&=\mathcal V_p^{(r)}+\mathcal M^{(r+1)}-\mathcal Z_p^{(r+1)},
&&p\in\mathcal I_{\mathrm{SP}},\\
\mathcal P_g^{(r+1)}
&=\mathcal P_g^{(r)}+\mathcal M^{(r+1)}-\mathcal G_g^{(r+1)},
&&g\in\mathcal I_{\mathrm{GEO}},\\
\mathcal D_{\mathrm{AR}}^{(r+1)}
&=\mathcal D_{\mathrm{AR}}^{(r)}+\mathcal M^{(r+1)}-\mathcal A^{(r+1)},
&&\text{if }\iota_{\mathrm{AR}}=1.
\end{aligned}
\label{eq:dual_updates}
\end{equation}

For a compact statement of the stopping rule, let $z^{(r)}\in\mathbb R^{qN}$ stack the vectorized active auxiliary tensors, let $u^{(r)}$ stack their scaled dual variables in the same order, and define $P\operatorname{vec}(\mathcal M)=\mathbf 1_q\otimes\operatorname{vec}(\mathcal M)$. We monitor the consensus residual
\begin{equation*}
r_{\mathrm{pri}}^{(r+1)}
=
\left\|
P\operatorname{vec}(\mathcal M^{(r+1)})-z^{(r+1)}
\right\|_2
\end{equation*}
and the blockwise dual-change diagnostic
\begin{equation*}
r_{\mathrm{dual}}^{(r+1)}
=
\beta\|z^{(r+1)}-z^{(r)}\|_2.
\end{equation*}
These definitions are equivalent to the sums of squared Frobenius norms over the active blocks used in implementation. Given absolute and relative tolerances $\epsilon_{\mathrm{abs}}$ and $\epsilon_{\mathrm{rel}}$, set
\begin{equation}
\begin{aligned}
\epsilon_{\mathrm{pri}}^{(r+1)}
&=
\sqrt{qN}\,\epsilon_{\mathrm{abs}}
+
\epsilon_{\mathrm{rel}}
\max\!\left\{
\|P\operatorname{vec}(\mathcal M^{(r+1)})\|_2,
\|z^{(r+1)}\|_2
\right\},\\
\epsilon_{\mathrm{dual}}^{(r+1)}
&=
\sqrt{qN}\,\epsilon_{\mathrm{abs}}
+
\epsilon_{\mathrm{rel}}\,\beta\|u^{(r+1)}\|_2.
\end{aligned}
\label{eq:stopping_tolerances}
\end{equation}
At a fixed value of $\beta$, the inner iterations terminate when $r_{\mathrm{pri}}^{(r+1)}\leq\epsilon_{\mathrm{pri}}^{(r+1)}$ and $r_{\mathrm{dual}}^{(r+1)}\leq\epsilon_{\mathrm{dual}}^{(r+1)}$. The first condition controls violation of the consensus constraints, whereas the second requires every active auxiliary block to stabilize. For fixed $\beta>0$ and $\phi$, \eqref{eq:consensus_problem} is a convex two-block consensus problem when $\mathscr S_s$ and the structural operators are positive semidefinite. Under the population-curvature condition used in Section~\ref{sec:theory}, the primal solution is unique; standard two-block ADMM theory then ensures convergence to a primal--dual solution.

To accelerate computation, we optionally increase the algorithmic parameter across a finite sequence of stages while keeping all statistical regularization parameters fixed. Starting from $\beta_0$, after the stopping conditions for the current stage are met, set
\[
\beta_{\mathrm{old}}\leftarrow\beta,
\qquad
\beta\leftarrow\min\{c_\beta\beta,\beta_{\max}\},
\qquad c_\beta>1.
\]
Whenever $\beta$ changes, the active scaled dual variables are multiplied by $\beta_{\mathrm{old}}/\beta$, which preserves the corresponding unscaled Lagrange multipliers, and the GEO and AR systems are refactorized. The final stage is run at fixed $\beta_{\max}$ until the stopping conditions are met.
When $|\mathcal S|>1$, neighboring targets generally have similar weights and solutions. We therefore order the target locations and use the converged primal, auxiliary, and dual variables from one target to initialize the next. This warm start changes only the computation, not the estimator at convergence. The dominant costs are the LR singular value decompositions and the primal linear-system solve. %
Algorithm~\ref{alg:hm_admm} summarizes the procedure.

\begin{algorithm}[!h]
\footnotesize
\caption{HM-ADMM for heterogeneous-mode tensor recovery}
\label{alg:hm_admm}
\renewcommand{\arraystretch}{1.08}
\begin{tabularx}{\dimexpr\linewidth-\algomargin\relax}{@{}r@{\quad}X@{}}
\textbf{Input:}
&
Observations and population design operator $\mathscr S_s$; $K$ and $h$ when $|\mathcal S|>1$; active structural sets and operators; fixed regularization parameters; $\beta_0$, $c_\beta$, and $\beta_{\max}$; $\epsilon_{\mathrm{abs}}$, $\epsilon_{\mathrm{rel}}$, and an iteration limit.
\\
\textbf{Output:}
&
$\{\widetilde{\mathcal M}(s):s\in\mathcal S\}$ when $|\mathcal S|>1$, or $\widetilde{\mathcal M}$ when $\mathcal S=\{s_0\}$.
\\
1.
&
Order the target locations when $|\mathcal S|>1$; otherwise consider the single recovery problem.
\\
2.
&
For the current target, form the kernel weights when applicable, compute $\widehat{\mathcal B}_s$, and initialize the primal, auxiliary, and scaled-dual variables, using the preceding target as a warm start when available.
\\
3.
&
Set $\beta=\beta_0$ and factorize the active GEO and AR systems.
\\
4.
&
At the current fixed value of $\beta$, repeat until the residual criteria in \eqref{eq:stopping_tolerances} are satisfied or the iteration limit is reached:
\\
&
(a) update $\mathcal M$ by \eqref{eq:M_linear_system};
\\
&
(b) apply the active structural updates \eqref{eq:Y_update}--\eqref{eq:A_update};
\\
&
(c) update the scaled dual variables by \eqref{eq:dual_updates} and evaluate the residuals.
\\
5.
&
If the iteration limit is reached before the residual criteria hold, terminate and report nonconvergence. Otherwise, if $\beta<\beta_{\max}$, increase $\beta$, rescale the active scaled dual variables, refactorize the GEO and AR systems, and return to Step 4.
\\
6.
&
Store the final primal tensor and proceed to the next target, if any.
\end{tabularx}
\end{algorithm}

\section{Recovery Guarantees}
\label{sec:theory}

This section develops recovery guarantees for heterogeneous regularization. The general theory separates the contributions of the decomposable LR and SP penalties and the quadratic GEO and AR penalties to recovery error. It also identifies conditions under which quadratic regularization sharpens the global Frobenius-error guarantee rather than only controlling penalized eigendirections, and quantifies the bias--variance trade-off induced by local borrowing. We first establish the results for general trace regression and then specialize them to regular smooth domains and uniform tensor completion.

Fix a target location $s\in\mathcal S$ and suppress its dependence throughout this section whenever no ambiguity arises. Write $\mathcal M^*=\mathcal M^*(s)$, $\widetilde{\mathcal M}=\widetilde{\mathcal M}(s)$, $\Delta=\widetilde{\mathcal M}-\mathcal M^*$, $\widetilde{L}=\widetilde{L}_s$, $\mathscr S=\mathscr S_s$, and $\widehat{\mathcal B}=\widehat{\mathcal B}_s$. When $\mathcal S=\{s_0\}$, the index is suppressed and these quantities describe ordinary tensor recovery. Dependence on $s$ is restored only when it is needed to describe local smoothing or the sampling weights.

\subsection{General recovery bounds}\label{subsec:squared_curvature}

Recovery in Frobenius norm first requires the population loss to
distinguish perturbations of the target tensor. Because
$\widetilde L$ is quadratic, this property is determined entirely by
the population second-moment operator $\mathscr S$. Indeed, $\nabla\widetilde{L}(\mathcal M)=2\mathscr S(\mathcal M)-2\widehat{\mathcal B}$. Hence, for every tensor $\mathcal H$ and $u\in[0,1]$, $\nabla\widetilde{L}(\mathcal M^*+u\mathcal H)-\nabla\widetilde{L}(\mathcal M^*)=2u\mathscr S(\mathcal H)$.

\begin{assumption}
\label{ass:population_curvature}
There exists $\mu>0$ such that $\langle\mathcal H,\mathscr S(\mathcal H)\rangle\geq\mu\|\mathcal H\|_F^2$ for every tensor $\mathcal H$.
\end{assumption}

Assumption~\ref{ass:population_curvature} rules out tensor directions
that are invisible to the population design. Equivalently, the matrix representation of \(\mathscr S\) has minimum eigenvalue bounded below by \(\mu\). Conditions of this type are standard in regularized trace regression and play the same role as strong or restricted strong convexity in analyses of decomposable regularization \citep{negahban2012unified,raskutti2019tensor,chen2024dynamic}. The assumption holds for isotropic designs and, more generally, whenever the population design covariance is uniformly nonsingular. It also holds under uniform tensor completion, where \(\mathscr S(\mathcal H)=\mathcal H/N\) and \(\mu=1/N\). In the locally weighted regime, the condition is preserved when the population design operators at all locations receiving positive weight share a common positive lower eigenvalue bound. Since \(\widetilde L\) is quadratic, Assumption~\ref{ass:population_curvature} yields the exact Bregman identity
\[
\widetilde{L}(\mathcal M^*+\mathcal H)-\widetilde{L}(\mathcal M^*)
-\langle\nabla\widetilde{L}(\mathcal M^*),\mathcal H\rangle
=\langle\mathcal H,\mathscr S(\mathcal H)\rangle
\geq\mu\|\mathcal H\|_F^2.
\]
Consequently, \(\widetilde L\) is globally \(2\mu\)-strongly convex with respect to the Frobenius norm.

Two quantities prevent $\mathcal M^*$ from being an exact stationary
point of the empirical criterion: random score fluctuation and, under
local averaging, deterministic population-score bias. To separate
their effects, define
\[
\mathcal Z
=
\nabla\widetilde{L}(\mathcal M^*)-
\mathbb E\nabla\widetilde{L}(\mathcal M^*)
=
-2\{\widehat{\mathcal B}-\mathbb E\widehat{\mathcal B}\},
\qquad
b=\|\mathbb E\nabla\widetilde{L}(\mathcal M^*)\|_F.
\]
In the correctly specified singleton-index regime, $b=0$. When $|\mathcal S|>1$, local averaging may shift the population score away from zero; the size of this deterministic bias is controlled below under smoothness conditions on the tensor field and regularity conditions on the local population design. Let $K_{\mathrm{LR}}=|\mathcal I_{\mathrm{LR}}|$ and $K_{\mathrm{SP}}=|\mathcal I_{\mathrm{SP}}|$. Set
\[
\xi
=
\begin{cases}
b, & K_{\mathrm{LR}}+K_{\mathrm{SP}}>0,\\[2mm]
b+\|\mathcal Z\|_F, & K_{\mathrm{LR}}=K_{\mathrm{SP}}=0.
\end{cases}
\]
For $l\in\mathcal I_{\mathrm{LR}}$, let $r_l=\operatorname{rank}(\mathcal M^*_{(l)})$. For $p\in\mathcal I_{\mathrm{SP}}$, let $S_p=\{j:[\mathcal M^*_{(p)}]_{j,:}\neq0\}$ and $s_p=|S_p|$. Define
\[
\Psi
=
\frac{1+\sqrt2}{2}\lambda_{\mathrm{LR}}
\sum_{l\in\mathcal I_{\mathrm{LR}}}\sqrt{r_l}
+
\lambda_{\mathrm{SP}}
\sum_{p\in\mathcal I_{\mathrm{SP}}}\sqrt{s_p}.
\]
The coefficients in $\Psi$ are determined by the decomposable geometry of the two degree-one penalties. For an LR mode, the tangent part of the score is controlled by $\sqrt{2r_l}\|\Delta\|_F$, while the canonical nuclear-norm subgradient contributes $\sqrt{r_l}\|\Delta\|_F$. For an SP mode, the active-row score term and the group-norm subgradient each contribute at most $\sqrt{s_p}\|\Delta\|_F$. Hence the total degree-one contribution in the first-order inequality is $2\Psi\|\Delta\|_F$. The quadratic structural energy is represented separately by $Q$. In the ordinary model $y_i=\langle\mathcal X_i,\mathcal M^*\rangle+\varepsilon_i$, the score decomposes as
\[
\mathcal Z
=
-2\{(\widehat{\mathscr S}-\mathscr S)(\mathcal M^*)\}
-
\frac{2}{m}\sum_{i=1}^m\varepsilon_i\mathcal X_i,
\]
where $\widehat{\mathscr S}(\mathcal H)=m^{-1}\sum_i\langle\mathcal X_i,\mathcal H\rangle\mathcal X_i$. Thus the score contains both signal--design and noise fluctuations.

To translate score fluctuation into rank- and support-dependent error
control, each active degree-one penalty must dominate the realized
centered score in its corresponding dual norm. For the LR and SP families, respectively, define $W_{\mathrm{LR}}=\max_{l\in\mathcal I_{\mathrm{LR}}}\|\mathcal Z_{(l)}\|_{\mathrm{op}}$, $W_{\mathrm{SP}}=\max_{p\in\mathcal I_{\mathrm{SP}}}\|\mathcal Z_{(p)}\|_{2,\infty}$, where $\|A\|_{2,\infty}=\max_j\|A_{j,:}\|_2$. We require the tuning parameters for the active degree-one families to satisfy
\begin{equation}
\lambda_{\mathrm{LR}}K_{\mathrm{LR}}
\geq
\frac{W_{\mathrm{LR}}}{1+\mathbf 1\{K_{\mathrm{SP}}>0\}},
\qquad
\lambda_{\mathrm{SP}}K_{\mathrm{SP}}
\geq
\frac{W_{\mathrm{SP}}}{1+\mathbf 1\{K_{\mathrm{LR}}>0\}},
\label{eq:score_domination}
\end{equation}
where $\mathbf 1\{\cdot\}$ is the indicator and the inequality for an inactive family is omitted. The denominators split the score equally between the LR and SP penalties when both families are active; otherwise, the sole active family controls the full score.%

\begin{theorem}
\label{thm:heterogeneous_deterministic_bound}
Consider the ordinary or locally weighted population-quadratic estimator. Suppose Assumption~\ref{ass:population_curvature} and the tuning inequalities in \eqref{eq:score_domination} hold for every active degree-one family. Then
\begin{equation}
\|\Delta\|_F
\leq
\min\left\{
\frac{
\xi+2\Psi+
\sqrt{(\xi+2\Psi)^2+4\mu Q(\mathcal M^*)}
}{4\mu},
\frac{\xi}{2\mu}
+
\sqrt{
\frac{\xi^2}{4\mu^2}
+
\frac{2R(\mathcal M^*)+Q(\mathcal M^*)}{\mu}
}
\right\}.
\label{eq:heterogeneous_frobenius_bound}
\end{equation}
Moreover,
\begin{equation}
Q(\Delta)
\leq
Q(\mathcal M^*)
+
\frac{(\xi+2\Psi)^2}{8\mu}.
\label{eq:explicit_quadratic_error_bound}
\end{equation}
\end{theorem}

Theorem~\ref{thm:heterogeneous_deterministic_bound} separates the roles
of the two types of regularization. The first branch exploits the
tangent spaces of the LR modes and the active rows of the SP modes,
leading to the rank and support factors in $\Psi$. The second branch
depends directly on $R(\mathcal M^*)$ and may be preferable when these
rank or support descriptions are loose. The GEO and AR penalties enter
through their mismatch with the truth, $Q(\mathcal M^*)$, while
\eqref{eq:explicit_quadratic_error_bound} directly controls the
quadratic structural error $Q(\Delta)$. 

Two distinct special cases should be distinguished. First, suppose that at least one quadratic penalty is active, so that \(Q(\mathcal M)\not\equiv0\), but the truth is exactly compatible with every active quadratic structure \(Q(\mathcal M^*)=0\). In this case, the GEO and AR penalties remain in the estimator, while \(\mathcal M^*\) lies in the joint null space of their structural operators. Theorem~\ref{thm:heterogeneous_deterministic_bound} then gives
\[
\|\Delta\|_F
\leq
\frac{\xi+2\Psi}{2\mu},
\qquad
Q(\Delta)
\leq
\frac{(\xi+2\Psi)^2}{8\mu}.
\]
Second, if no quadratic penalty is imposed, then \(Q(\mathcal M)\equiv0\) for every \(\mathcal M\), and the estimator contains only the active LR and SP penalties. The corresponding joint LR+SP, LR-only, and SP-only bounds are given in Corollary~\textcolor{black}{S.1} of the Supplementary Material.

To interpret the error control provided by \(Q(\Delta)\), we decompose \(\Delta\) along the eigenspaces of the GEO and AR operators: directions with small eigenvalues are graph-smooth or have low AR discrepancy, whereas those with large eigenvalues represent stronger departures from the imposed structure.

\begin{corollary}%
\label{cor:spectral_quadratic_control}
For $g\in\mathcal I_{\mathrm{GEO}}$, let $0\leq\nu_{g,1}\leq\cdots\leq\nu_{g,n_g}$ be the eigenvalues of $L_g$, and let $P_{g,r}^{\perp}$ denote the orthogonal projection onto the eigenspaces associated with $\nu_{g,r+1},\ldots,\nu_{g,n_g}$. Whenever $\nu_{g,r+1}>0$,
\[
\|P_{g,r}^{\perp}\Delta_{(g)}\|_F^2
\leq
\frac{
Q(\mathcal M^*)+(\xi+2\Psi)^2/(8\mu)
}{
\lambda_{\mathrm{GEO}}\nu_{g,r+1}
}.
\]
Likewise, let $0\leq\omega_1\leq\cdots\leq\omega_T$ be the eigenvalues of $D_\phi^\top D_\phi$, and let $P_{\mathrm{AR},r}^{\perp}$ denote the orthogonal projection onto the eigenspaces associated with $\omega_{r+1},\ldots,\omega_T$. When $\omega_{r+1}>0$,
\[
\|P_{\mathrm{AR},r}^{\perp}\Delta_{(a)}\|_F^2
\leq
\frac{
Q(\mathcal M^*)+(\xi+2\Psi)^2/(8\mu)
}{
\lambda_{\mathrm{AR}}\omega_{r+1}
}.
\]

\end{corollary}

Because graph Laplacians and AR difference operators may have nontrivial null spaces, the bounds control eigenspaces separated from zero by a positive spectral gap. They distinguish the structural mismatch \(Q(\mathcal M^*)\) from the score and LR/SP complexity term \((\xi+2\Psi)^2/(8\mu)\). Corollary~\textcolor{black}{S.2} in the Supplementary Material gives a general coercive-subspace extension.

Theorem~\ref{thm:heterogeneous_deterministic_bound} is a pathwise result requiring the active degree-one penalties to dominate the realized score. We next control this score under independent trace regression. Here, for each target \(s\), independence means that the observation pairs \(\{(\mathcal X_{s'i},\varepsilon_{s'i}):(s',i)\in\mathcal J_s\}\) entering the local criterion are mutually independent, although their distributions may vary across locations. Under standard tail conditions, this independence yields explicit high-probability control of the score. The observation weights enter through $m_{\mathrm{eff}}=(\sum_j a_j^2)^{-1}$ and $a_{\max}=\max_j a_j$.  For every $t\geq1$, let $\omega_{\mathrm{LR}}(t)$, $\omega_{\mathrm{SP}}(t)$, $\omega_F(t)$ be deterministic score radii.  When at least one degree-one family is active, define the relevant score event by $\mathcal E_t=\left\{W_{\mathrm{LR}}\leq\omega_{\mathrm{LR}}(t),W_{\mathrm{SP}}\leq\omega_{\mathrm{SP}}(t)\right\}$, where the inequality for an inactive family is omitted.  When both degree-one families are absent, set $\mathcal E_t=\{\|\mathcal Z\|_F\leq\omega_F(t)\}$. Choose the radii so that
\begin{equation}
\mathbb P(\mathcal E_t)\geq1-2e^{-t}.
\label{eq:score_radius_event}
\end{equation}
Proposition~\textcolor{black}{S.1} in the Supplementary Material provides explicit Bernstein choices of these radii.

For the statistical bound, choose
\[
\lambda_{\mathrm{LR}}
=
\begin{cases}
\omega_{\mathrm{LR}}(t)/(2K_{\mathrm{LR}}), & K_{\mathrm{SP}}>0,\\
\omega_{\mathrm{LR}}(t)/K_{\mathrm{LR}}, & K_{\mathrm{SP}}=0,
\end{cases}
\qquad
\lambda_{\mathrm{SP}}
=
\begin{cases}
\omega_{\mathrm{SP}}(t)/(2K_{\mathrm{SP}}), & K_{\mathrm{LR}}>0,\\
\omega_{\mathrm{SP}}(t)/K_{\mathrm{SP}}, & K_{\mathrm{LR}}=0,
\end{cases}
\]
with absent families omitted. Let $\Psi_t$ denote $\Psi$ evaluated at these tuning levels, and set
\[
\xi_t
=
\begin{cases}
b, & K_{\mathrm{LR}}+K_{\mathrm{SP}}>0,\\
b+\omega_F(t), & K_{\mathrm{LR}}=K_{\mathrm{SP}}=0.
\end{cases}
\]
Finally, define the aggregate score--complexity radius
\begin{equation*}
\eta_t=\xi_t+2\Psi_t.
\end{equation*}

\begin{theorem}%
\label{thm:independent_trace_recovery}
Suppose Assumption~\ref{ass:population_curvature} holds. For any score radii satisfying \eqref{eq:score_radius_event}, use the tuning rule above. Then, for every $t\geq1$, with probability at least $1-2e^{-t}$,
\[
\|\Delta\|_F
\leq
\min\left\{
\frac{\eta_t+\sqrt{\eta_t^2+4\mu Q(\mathcal M^*)}}{4\mu},
\frac{\xi_t}{2\mu}
+
\sqrt{
\frac{\xi_t^2}{4\mu^2}
+
\frac{2R(\mathcal M^*)+Q(\mathcal M^*)}{\mu}}
\right\},
\]
and
\[
Q(\Delta)
\leq
Q(\mathcal M^*)+
\frac{\eta_t^2}{8\mu}.
\]
\end{theorem}

Theorem~\ref{thm:independent_trace_recovery} converts the pathwise
oracle inequality into a statistical guarantee at confidence level
$1-2e^{-t}$. Its radius separates smoothing bias $b$, random score
fluctuation through the score radii, rank and support complexity through
$\Psi_t$, and mismatch with the imposed quadratic structures through
$Q(\mathcal M^*)$. The result does not by itself show that the GEO or
AR penalties improve the global Frobenius-error bound, however, because
their quadratic operators may have nontrivial null spaces. We next
identify conditions under which such an improvement can be established.

To isolate the contribution of quadratic regularization, we use as a
benchmark the otherwise identical estimator with the same LR and SP
penalties and tuning levels but with $Q=0$. A global comparison is
possible when the admissible error directions lie in a deterministic
subspace $\mathbb H$ on which $Q$ is coercive. Here, coercivity means
that the quadratic penalty uniformly controls the Frobenius norm over
$\mathbb H$, as formalized by the condition $\tau_{\mathbb H}>0$
below.

\begin{corollary}%
\label{cor:quadratic_structure_gain}
Suppose the conditions of Theorem~\ref{thm:independent_trace_recovery} hold, and let $\Delta_Q$ denote the error of the estimator with the quadratic penalty $Q$. Suppose that $\Delta_Q\in\mathbb H$ almost surely for a deterministic linear subspace $\mathbb H$ and $\tau_{\mathbb H} = \inf_{\mathcal H\in\mathbb H,\|\mathcal H\|_F=1} Q(\mathcal H) >0$.
Then, for every $t\geq1$, with probability at least $1-2e^{-t}$,
\[
\|\Delta_Q\|_F
\leq
B_{Q,t}
=
\frac{
\eta_t+
\sqrt{\eta_t^2+4(2\mu+\tau_{\mathbb H})Q(\mathcal M^*)}
}{
2(2\mu+\tau_{\mathbb H})
}.
\]
The corresponding rank--support branch for the estimator obtained by setting $Q=0$, with the same degree-one penalties and tuning levels, is $B_{0,t}=\eta_t/(2\mu)$. For $\eta_t>0$, $B_{Q,t}<B_{0,t}$ whenever \( Q(\mathcal M^*) <{\tau_{\mathbb H}\eta_t^2}/{(4\mu^2)}. \) In particular, if $\eta_t>0$, exact compatibility yields
\[
B_{Q,t}
=
\frac{\eta_t}{2\mu+\tau_{\mathbb H}}
<
\frac{\eta_t}{2\mu}
=
B_{0,t}.
\]
\end{corollary}

Corollary~\ref{cor:quadratic_structure_gain} makes the benefit--mismatch
trade-off explicit. The quantity $\tau_{\mathbb H}$ measures the
additional curvature supplied by the quadratic penalties, whereas
$Q(\mathcal M^*)$ measures their mismatch with the truth. The guaranteed
radius is strictly smaller when the latter is below the stated
threshold; under exact compatibility, the improvement factor is
$2\mu/(2\mu+\tau_{\mathbb H})$. This is a comparison of guaranteed
error radii, not pointwise dominance of one estimator over another.
If coercivity fails, the spectral bounds still control penalized
eigendirections, but no global ordering follows. When
$\mathcal S=\{s_0\}$, the same comparison isolates the contribution of
the retained-mode structures without relying on local smoothing.

\subsection{Local smoothing and optimal bandwidth}
\label{subsec:smooth_bandwidth}

The general bounds do not yet determine whether local borrowing is beneficial, because both its smoothing bias and its stochastic error depend on the bandwidth. We now impose regular smooth-domain conditions under which these two contributions can be quantified and balanced.

\begin{assumption}%
\label{ass:regular_smooth_kernel}
Let $d\geq1$, let $|\mathcal S|>1$, and let $s$ be an interior target. Throughout the local window around $s$, the population design operators are identical, $\mathscr S_{s'}=\mathscr S$, and satisfy $\mu I\preceq\mathscr S\preceq\bar\mu I$. The kernel $K$ is nonnegative, symmetric, bounded, and supported on $\{u\in\mathbb R^d:\|u\|_2\leq1\}$. Moreover, $\int K(u)\,du=1$ and $\alpha(K)=\int\|u\|_2^2K(u)\,du<\infty$.
\end{assumption}

\begin{assumption}%
\label{ass:smooth_tensor_field}
The tensor field $\mathcal M^*(\cdot)$ is twice continuously differentiable in Frobenius norm on the relevant neighborhood, with \(D_2=\sup_{u,\,v:\|v\|_2=1} \|\partial_v^2\mathcal M^*(u)\|_F<\infty. \)
\end{assumption}

\begin{assumption}%
\label{ass:regular_fixed_design}
Let $L$ be the total number of design locations. Along the asymptotic sequence, $h\to0$ and $Lh^d\to\infty$. The kernel window around $s$ contains $\asymp Lh^d$ design locations, $m_{s'}\asymp m$ uniformly within the window, and \( \max_{s'}w_h(s',s)\lesssim(Lh^d)^{-1}. \) Defining $\overline{\mathcal M}_h^*(s)=\sum_{s'}w_h(s',s)\mathcal M^*(s')$, assume further that
\[
\left\|
\overline{\mathcal M}_h^*(s)
-
\int K(u)\mathcal M^*(s+hu)\,du
\right\|_F=o(h^2).
\]
\end{assumption}

The counting and weight conditions in Assumption~\ref{ass:regular_fixed_design} hold for quasi-uniform fixed designs whose local design density is bounded above and away from zero. The final condition requires the kernel quadrature error to be negligible relative to the leading smoothing bias and is satisfied when the design grid is sufficiently dense relative to the bandwidth. Because $a_{s,(s',i)}=w_h(s',s)/m_{s'}$, this assumption implies
\begin{equation}
m_{\mathrm{eff}}\asymp mLh^d,
\qquad
a_{\max}\asymp(mLh^d)^{-1}.
\label{eq:regular_fixed_design_scaling}
\end{equation}

Symmetry of the kernel eliminates the first-order Taylor term. The argument in Proposition~\textcolor{black}{S.2} therefore gives
\begin{equation}
b(h)\leq\bar\mu\,\alpha(K)D_2h^2+o(h^2).
\label{eq:smooth_bias}
\end{equation}
For the Bernstein score radii underlying Theorem~\ref{thm:independent_trace_recovery}, the fixed-design scaling in \eqref{eq:regular_fixed_design_scaling} similarly yields
\begin{equation}
\eta_t(h)
\leq
b(h)
+\frac{C\Lambda_t}{\sqrt{mLh^d}}
+\frac{C\Theta_t}{mLh^d},
\label{eq:smooth_score_radius}
\end{equation}
uniformly over the bandwidths under consideration. The explicit variance--complexity coefficient $\Lambda_t$ and Bernstein-envelope coefficient $\Theta_t$ are given in \textcolor{black}{(S.3.2)} of the Supplementary Material. We consider the regime $\Theta_t=o\{\Lambda_t\sqrt{mLh^d}\}$, in which the last term in \eqref{eq:smooth_score_radius} is of smaller order.

Write $D_K=\alpha(K)D_2$ and define
\[
F_q(u)=\frac{u+\sqrt{u^2+4\mu q}}{4\mu},
\qquad q,u\geq0.
\]
The leading bandwidth-dependent terms are then $\bar\mu D_Kh^2$ and $C\Lambda_t(mLh^d)^{-1/2}$. The former is the local smoothing bias, whereas the latter is the stochastic error after pooling approximately $mLh^d$ observations. Their balance gives the familiar order $h\asymp(mL)^{-1/(d+4)}$ after suppressing structural, design, and logarithmic factors.

For an interior target $s$, let
\begin{equation*}
q_s=Q\{\mathcal M^*(s)\}
=
\lambda_{\mathrm{GEO}}
\sum_{g\in\mathcal I_{\mathrm{GEO}}}
\|\mathcal E_g^{\mathrm{GEO}}\{\mathcal M^*(s)\}\|_F^2
+
\lambda_{\mathrm{AR}}
\sum_{u=2}^T
\|\mathcal M_u^*(s)-\phi\mathcal M_{u-1}^*(s)\|_F^2.
\end{equation*}
Either sum is omitted when the corresponding penalty is absent. This quantity measures departure from the active GEO and AR structures
and is not assumed to be small. Its role differs from that of the two
smoothing terms: for fixed quadratic tuning parameters, $q_s$ affects
the attainable error bound but is independent of $h$, and therefore
does not alter the bandwidth that balances smoothing bias and
stochastic variation.

\begin{theorem}
\label{thm:optimal_bandwidth_rate}
Fix $t\geq1$. Suppose Assumptions~\ref{ass:population_curvature}--\ref{ass:regular_fixed_design} hold with fixed positive $\mu,\bar\mu,D_K$, and use the uniform Bernstein score bound \eqref{eq:smooth_score_radius}. If $\Lambda_t>0$, $\Theta_t=o\{\Lambda_t\sqrt{mLh^d}\}$ uniformly over the candidate range, and the deterministic minimizer of the leading bound lies in that range, then it satisfies
\begin{equation}
h_{\mathrm{opt}}
\asymp
\left\{
\frac{\Lambda_t^2}
{\bar\mu^2D_K^2mL}
\right\}^{1/(d+4)}.
\label{eq:optimal_bandwidth_general}
\end{equation}
Define
\[
\mathcal R_{t,\mathrm{opt}}
=
C_d\bar\mu D_K
\left\{
\frac{\Lambda_t^2}
{\bar\mu^2D_K^2mL}
\right\}^{2/(d+4)},
\]
where $C_d>0$ depends only on $d$ and the regular fixed-design constants. Then, with probability at least $1-2e^{-t}$,
\begin{equation}
\|\widetilde{\mathcal M}(s)-\mathcal M^*(s)\|_F
\leq
F_{q_s}(\mathcal R_{t,\mathrm{opt}})
+o(h_{\mathrm{opt}}^2).
\label{eq:optimal_rate_general}
\end{equation}
\end{theorem}

The completion bound preserves the same separation of effects: local
pooling acts through $b$ and $m_{\mathrm{eff}}$, LR and SP complexity
through $\Lambda_t^{\mathrm{cp}}$, and quadratic mismatch through
$Q(\mathcal M^*)$. 
Under exact compatibility, $q_s=0$ and $F_0(u)=u/(2\mu)$, so the bandwidth-dependent recovery rate has order $(mL)^{-2/(d+4)}$, up to the factors collected in $\Lambda_t$. More generally, the mismatch does not change this order when $q_s=O(\{\mathcal R_{t,\mathrm{opt}}\}^2/\mu)$ and is negligible when the relation is $o(\{\mathcal R_{t,\mathrm{opt}}\}^2/\mu)$. A fixed positive $q_s$ instead leaves a residual term of order $\sqrt{q_s/\mu}$ in the bound, although the order of $h_{\mathrm{opt}}$ remains unchanged. The graph-spectral and autoregressive innovation models in the Supplementary Material provide probabilistic upper bounds for $q_s$ under additional assumptions.

\subsection{Uniform-completion specializations}
\label{subsec:squared_completion}

We now specialize the preceding results to uniform tensor completion. Suppose each design tensor is a canonical basis tensor $\mathcal E_{\xi_j}$, where $\xi_j$ is sampled independently and uniformly from the $N=\prod_{r=1}^k n_r$ entries of the retained tensor. Then
\[
\mathscr S(\mathcal H)=\frac{\mathcal H}{N},
\qquad
\mu=\bar\mu=\frac1N.
\]
The Frobenius loss $\|\Delta\|_F/\sqrt N$ is therefore the root mean squared entrywise recovery error. In the locally weighted regime, $b=2N^{-1}
\|\mathcal M^*(s)-\overline{\mathcal M}_h^*(s)\|_F$, and with $D_K=\alpha(K)D_2$, Assumptions~\ref{ass:regular_smooth_kernel}--\ref{ass:regular_fixed_design} give
\begin{equation}
b\leq\frac{D_Kh^2}{N}+o(h^2/N).
\label{eq:completion_bias}
\end{equation}
In the correctly specified singleton-index regime, $b=0$.

Assume that $y_j=\mathcal M_j^*(\xi_j)+\varepsilon_j$, $\mathbb E(\varepsilon_j\mid\xi_j)=0$, $\|\varepsilon_j\|_{\psi_2}\leq\sigma$, and $B=\sup_j\|\mathcal M_j^*\|_\infty<\infty$. Put $V=(B^2+\sigma^2)^{1/2}$ and $d_l=n_l\wedge(N/n_l)$ for $l\in\mathcal I_{\mathrm{LR}}$. General completion score bounds are given in the Supplementary Material; for a concise main-text rate, we use a common-variance specialization.

\begin{assumption}%
\label{ass:nondegenerate_completion_noise}
There exists a constant $\sigma_\varepsilon^2>0$ such that $\mathbb E(\varepsilon_j^2\mid\xi_j)=\sigma_\varepsilon^2$ almost surely for every observation in the local window.
\end{assumption}

This assumption is used only to simplify the score scales. Suppose $a_{\max}\asymp m_{\mathrm{eff}}^{-1}$ and the sufficient effective-sample-size condition in \textcolor{black}{(S.4.3)} holds, so that the linear Bernstein terms are of smaller order. The coefficient $\Lambda_t^{\mathrm{cp}}$ in \textcolor{black}{(S.4.2)} then gives
\begin{equation}
\eta_t^{\mathrm{cp}}
\leq
b+\frac{C\Lambda_t^{\mathrm{cp}}}{\sqrt{m_{\mathrm{eff}}}}.
\label{eq:completion_score_radius}
\end{equation}
For $q,u\geq0$, define
\[
G_{N,q}(u)
=
\frac{\sqrt N}{4}
\left\{u+\sqrt{u^2+\frac{4q}{N}}\right\}.
\]

\begin{corollary}%
\label{cor:nondegenerate_completion}
Under Assumption~\ref{ass:nondegenerate_completion_noise} and the preceding score conditions, for every $t\geq1$, with probability at least $1-2e^{-t}$,
\begin{equation}
\frac{\|\Delta\|_F}{\sqrt N}
\leq
G_{N,Q(\mathcal M^*)}
\left(
b+\frac{C\Lambda_t^{\mathrm{cp}}}{\sqrt{m_{\mathrm{eff}}}}
\right).
\label{eq:completion_recovery}
\end{equation}
\end{corollary}

Under exact compatibility, $Q(\mathcal M^*)=0$ and $G_{N,0}(u)=\sqrt N u/2$, so \eqref{eq:completion_recovery} reduces to
\[
\frac{\|\Delta\|_F}{\sqrt N}
\leq
\frac{\sqrt N}{2}
\left(
b+\frac{C\Lambda_t^{\mathrm{cp}}}{\sqrt{m_{\mathrm{eff}}}}
\right).
\]
In the smooth-domain regime, \eqref{eq:completion_bias} and $m_{\mathrm{eff}}\asymp mLh^d$ yield the corresponding oracle bandwidth.

\begin{corollary}%
\label{cor:completion_optimal_bandwidth_rate}
Fix $t\geq1$. Under Corollary~\ref{cor:nondegenerate_completion} and Assumptions~\ref{ass:regular_smooth_kernel}--\ref{ass:regular_fixed_design}, suppose $D_K,\Lambda_t^{\mathrm{cp}}>0$ and \textcolor{black}{(S.4.3)} holds uniformly over the candidate range. If this range contains the deterministic minimizer of the leading bound, then the minimizer satisfies
\begin{equation}
h_{\mathrm{opt}}^{\mathrm{cp}}
\asymp
\left\{
\frac{N^2(\Lambda_t^{\mathrm{cp}})^2}
{D_K^2mL}
\right\}^{1/(d+4)}.
\label{eq:completion_optimal_bandwidth}
\end{equation}
Define
\begin{equation*}
\mathcal R_{t,\mathrm{opt}}^{\mathrm{cp}}
=
\frac{C_dD_K}{N}
\left\{
\frac{N^2(\Lambda_t^{\mathrm{cp}})^2}
{D_K^2mL}
\right\}^{2/(d+4)}.
\end{equation*}
Then, with probability at least $1-2e^{-t}$,
\begin{equation}
\frac{\|\widetilde{\mathcal M}(s)-\mathcal M^*(s)\|_F}{\sqrt N}
\leq
G_{N,Q\{\mathcal M^*(s)\}}
\bigl(\mathcal R_{t,\mathrm{opt}}^{\mathrm{cp}}\bigr)
+o\!\left\{
\frac{(h_{\mathrm{opt}}^{\mathrm{cp}})^2}{\sqrt N}
\right\}.
\label{eq:completion_optimal_rate}
\end{equation}
\end{corollary}

For fixed quadratic tuning parameters, $Q\{\mathcal M^*(s)\}$ is independent of $h$ and therefore does not change the bandwidth order. Its effect is governed by the near-compatibility conditions in Section~\ref{subsec:smooth_bandwidth}; exact compatibility follows by setting it to zero in \eqref{eq:completion_optimal_rate}.

\section{Simulation Studies}
\label{sec:simulation}

We conduct two simulation experiments to examine heterogeneous regularization with and without local borrowing. The first combines one smoothly varying indexing mode with all six pairwise combinations of the four retained-mode structures---LR, SP, GEO, and AR. Varying the observation rate assesses full-tensor recovery, while slice-wise errors show how performance changes along the smooth domain and near its boundaries. The second considers a singleton indexing domain with simultaneous LR, SP, and AR structure, thereby isolating the effect of heterogeneous regularization when no local borrowing is available. Recovery is evaluated by mean squared error, and the observation rate is denoted by $\rho$.

\subsection{Smooth tensor designs}
\label{subsec:sim_smooth}

We consider all pairwise combinations of the four heterogeneous mode structures. In each design, the generated tensor has dimension $20 \times 30 \times 40 \times 100$, with the fourth mode varying smoothly over $100$ indices. Details of the tensor generation procedures are provided in Section \textcolor{black}{S.6} of the Supplementary Material. Independent Gaussian noise with standard deviation one is added to the observed entries. For each combination and each $\rho \in \{0.05, 0.10, \ldots, 0.50\}$, we generate $100$ independent replications. The regularization parameters $\lambda_{\mathrm{LR}}$, $\lambda_{\mathrm{SP}}$, $\lambda_{\mathrm{GEO}}$, $\lambda_{\mathrm{AR}}$ and the bandwidth $h$ for HM-ADMM are selected by cross-validation and held fixed across the evaluation replications. For AR designs, $\phi$ is estimated from adjacent observed pairs using the fitting data only and held fixed during optimization as detailed in Supplementary Section~\textcolor{black}{S.6.3}.

The competitors are Gandy ADM-TR~\citep{gandy2011tensor} applied either to the full tensor or separately to slices of the smooth mode; we refer to these versions as Gandy-full and Gandy-slicewise. We additionally include Tian MNTC~\citep{tian2026manifold}, denoted Tian-MNTC, in the LR+GEO design and Takeuchi ATF~\citep{takeuchi2017autoregressive}, denoted Takeuchi-ATF, in the LR+AR design, where their structure-specific regularization is applicable. Thus, the competitor set is tailored to each design rather than padded with methods that do not encode the relevant structure. Recovery is measured by $\operatorname{MSE}(\widehat{\mathcal T})  =(\prod_{j=1}^4 n_j)^{-1}{\|\widehat{\mathcal T}-\mathcal T^*\|_F^2}$.

\begin{figure}[!h]
  \centering
  \includegraphics[width=\textwidth]{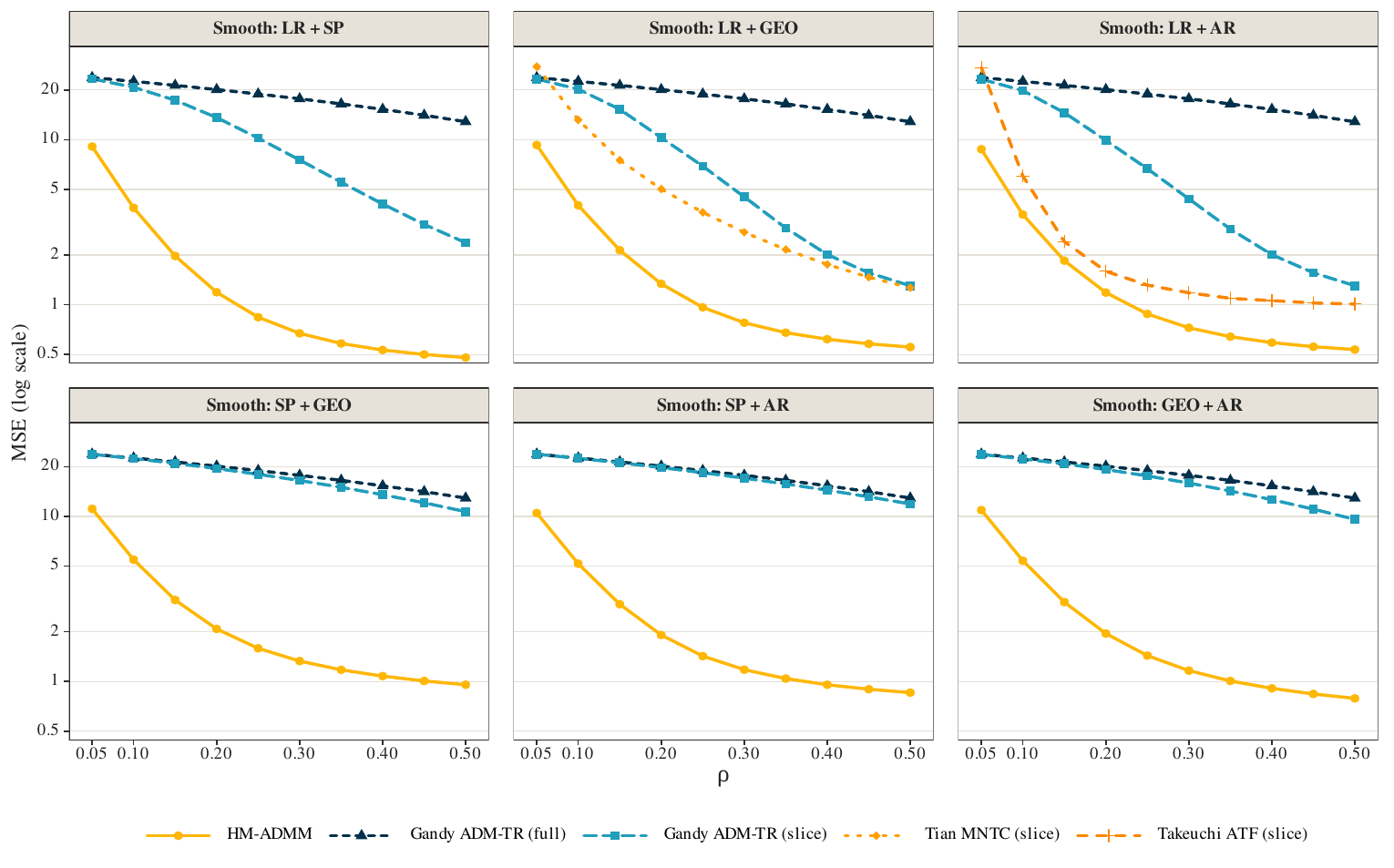}
  \caption{Mean MSE against the observation rate for six smooth tensor designs.  Curves are averages over 100 independent replications.}
  \label{fig:sim_smooth_rho}
\end{figure}

Figure~\ref{fig:sim_smooth_rho} and Table~\ref{tab:sim_smooth_mse} report mean MSE over the 100 replications.  All methods improve as the observation rate increases, but HM-ADMM has the lowest mean MSE in each of the 60 design--observation-rate combinations.  The gains are especially pronounced in the designs involving SP and in GEO+AR, and remain clear in LR+GEO and LR+AR.

\begin{table}[!h]
\centering
\caption{Mean full-tensor MSE for the six smooth tensor designs.  Each entry is averaged over 100 independent replications.}
\label{tab:sim_smooth_mse}
\fontsize{8}{9.6}\selectfont
\setlength{\tabcolsep}{1.8pt}
\begin{tabular}{@{}llrrrrrrrrrr@{}}
\toprule
Structure & Method & \multicolumn{10}{c}{$\rho$} \\
\cmidrule(lr){3-12}
 & & 0.05 & 0.10 & 0.15 & 0.20 & 0.25 & 0.30 & 0.35 & 0.40 & 0.45 & 0.50 \\
\midrule
LR+SP
 & Gandy-full      & 23.7813 & 22.5604 & 21.3393 & 20.1251 & 18.9090 & 17.6933 & 16.4823 & 15.2774 & 14.0663 & 12.8549 \\
 & Gandy-slicewise & 23.3678 & 20.7706 & 17.3417 & 13.6140 & 10.2310 &  7.5367 &  5.5288 &  4.0804 &  3.0697 &  2.3784 \\
 & \textbf{HM-ADMM} & \textbf{9.0868} & \textbf{3.8650} & \textbf{1.9754} & \textbf{1.1912} & \textbf{0.8412} & \textbf{0.6720} & \textbf{0.5832} & \textbf{0.5324} & \textbf{0.5005} & \textbf{0.4790} \\
\addlinespace
LR+GEO
 & Gandy-full      & 23.7773 & 22.5553 & 21.3366 & 20.1191 & 18.9024 & 17.6899 & 16.4772 & 15.2567 & 14.0480 & 12.8425 \\
 & Gandy-slicewise & 23.3072 & 20.2259 & 15.2776 & 10.2949 &  6.9134 &  4.5038 &  2.9172 &  2.0157 &  1.5634 &  1.3051 \\
 & Tian-MNTC       & 27.7325 & 13.1865 &  7.5159 &  5.0254 &  3.6287 &  2.7490 &  2.1617 &  1.7558 &  1.4694 &  1.2687 \\
 & \textbf{HM-ADMM} & \textbf{9.3092} & \textbf{3.9775} & \textbf{2.1257} & \textbf{1.3394} & \textbf{0.9716} & \textbf{0.7882} & \textbf{0.6879} & \textbf{0.6288} & \textbf{0.5903} & \textbf{0.5643} \\
\addlinespace
LR+AR
 & Gandy-full      & 23.7795 & 22.5559 & 21.3343 & 20.1175 & 18.9013 & 17.6888 & 16.4691 & 15.2577 & 14.0498 & 12.8376 \\
 & Gandy-slicewise & 23.2471 & 19.8445 & 14.5785 &  9.9021 &  6.6737 &  4.3807 &  2.8747 &  2.0121 &  1.5638 &  1.3047 \\
 & Takeuchi-ATF    & 27.3198 &  6.0380 &  2.4138 &  1.5964 &  1.3155 &  1.1825 &  1.0946 &  1.0601 &  1.0259 &  1.0104 \\
 & \textbf{HM-ADMM} & \textbf{8.7518} & \textbf{3.5205} & \textbf{1.8501} & \textbf{1.1863} & \textbf{0.8804} & \textbf{0.7260} & \textbf{0.6412} & \textbf{0.5905} & \textbf{0.5576} & \textbf{0.5356} \\
\addlinespace
SP+GEO
 & Gandy-full      & 23.7847 & 22.5692 & 21.3575 & 20.1494 & 18.9402 & 17.7330 & 16.5241 & 15.3175 & 14.1135 & 12.9097 \\
 & Gandy-slicewise & 23.7560 & 22.3880 & 20.9467 & 19.4728 & 17.9814 & 16.4892 & 15.0007 & 13.5303 & 12.0868 & 10.6785 \\
 & \textbf{HM-ADMM} & \textbf{11.1058} & \textbf{5.4660} & \textbf{3.1137} & \textbf{2.0803} & \textbf{1.5880} & \textbf{1.3291} & \textbf{1.1764} & \textbf{1.0775} & \textbf{1.0074} & \textbf{0.9556} \\
\addlinespace
SP+AR
 & Gandy-full      & 23.7887 & 22.5719 & 21.3585 & 20.1482 & 18.9394 & 17.7317 & 16.5260 & 15.3220 & 14.1166 & 12.9137 \\
 & Gandy-slicewise & 23.7805 & 22.4698 & 21.1128 & 19.7456 & 18.3832 & 17.0361 & 15.7100 & 14.4085 & 13.1311 & 11.8858 \\
 & \textbf{HM-ADMM} & \textbf{10.4759} & \textbf{5.1673} & \textbf{2.9366} & \textbf{1.9117} & \textbf{1.4252} & \textbf{1.1789} & \textbf{1.0412} & \textbf{0.9555} & \textbf{0.8977} & \textbf{0.8560} \\
\addlinespace
GEO+AR
 & Gandy-full      & 23.7832 & 22.5669 & 21.3523 & 20.1402 & 18.9286 & 17.7195 & 16.5116 & 15.3054 & 14.1002 & 12.8964 \\
 & Gandy-slicewise & 23.7124 & 22.2987 & 20.7979 & 19.2193 & 17.5800 & 15.9118 & 14.2483 & 12.6225 & 11.0609 &  9.5855 \\
& \textbf{HM-ADMM} & \textbf{10.9165} & \textbf{5.3884} & \textbf{3.0248} & \textbf{1.9514} & \textbf{1.4340} & \textbf{1.1635} & \textbf{1.0074} & \textbf{0.9084} & \textbf{0.8401} & \textbf{0.7899} \\
\bottomrule
\end{tabular}
\end{table}

\begin{figure}[!h]
  \centering
  \includegraphics[width=\textwidth,height=3in]{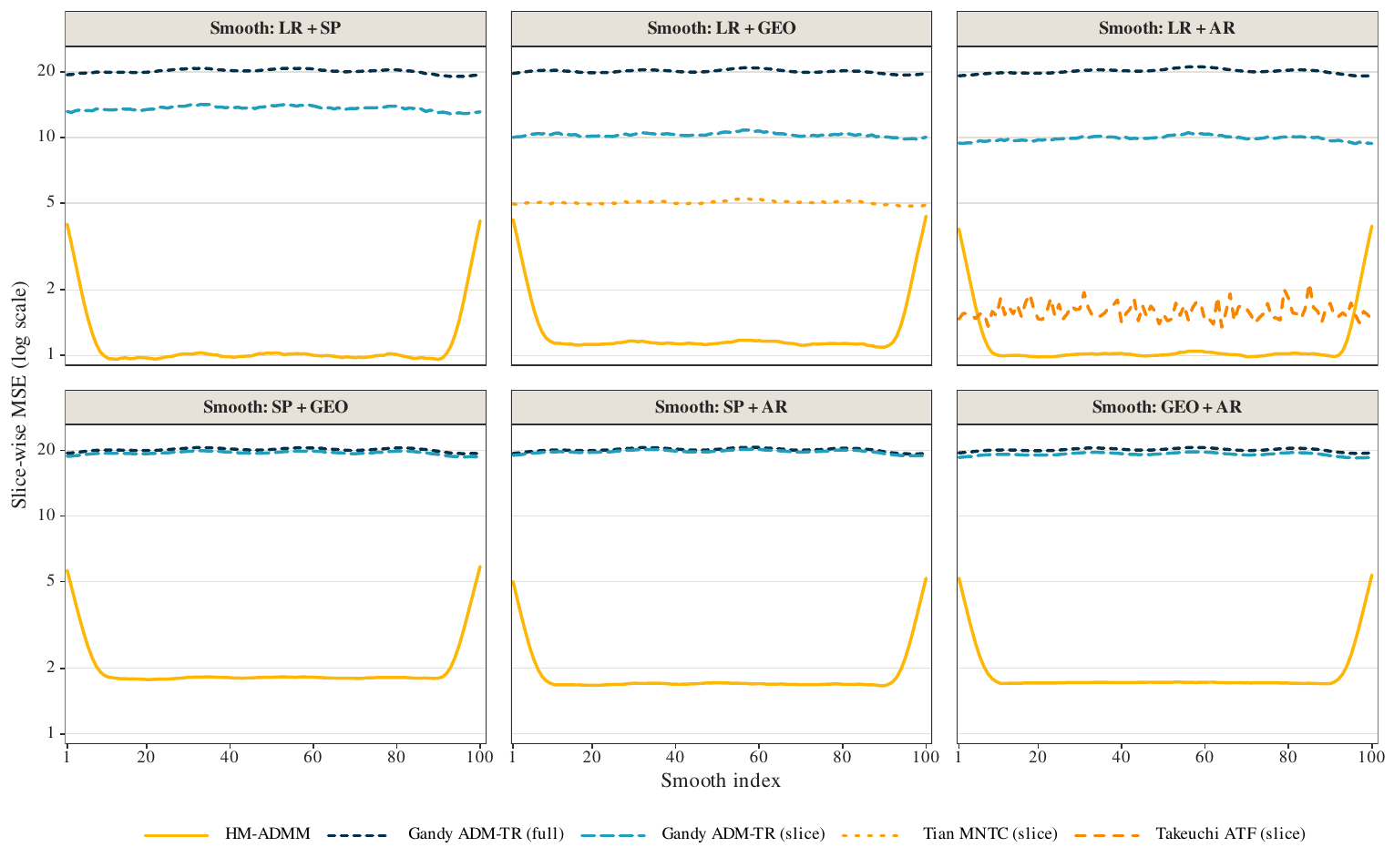}
  \caption{Mean slice-wise MSE  along the smooth mode at $\rho=0.20$.  Curves are averages over 100 independent replications and do not display confidence bands.  HM-ADMM exhibits a boundary effect, most visibly in the LR+AR design, where Takeuchi-ATF has lower error at the first and last five indices.}
  \label{fig:sim_smooth_slice}
\end{figure}

To determine whether these aggregate gains are confined to a small portion of the smooth domain, we next fix $\rho=0.20$ and compute slicewise MSE:
\[
  \operatorname{MSE}_s
  =\frac{1}{20\cdot30\cdot40}
  \bigl\|\widehat{\mathcal T}_{:,:,:,s}
  -\mathcal T^*_{:,:,:,s}\bigr\|_F^2,
  \qquad s=1,\ldots,100.
\]
Figure~\ref{fig:sim_smooth_slice} shows that HM-ADMM has lower mean slice-wise MSE at every index in five of the six designs.  In LR+AR, it performs best in the interior, whereas Takeuchi-ATF is better near the two endpoints.  The rise in HM-ADMM error near both ends is consistent with the reduced amount of two-sided local information at the boundary.

\subsection{A singleton-index LR+SP+AR design}
\label{subsec:sim_nosmooth}

We finally consider the degenerate case $\mathcal S=\{s_0\}$, so the indexing domain contains no neighboring location and $\mathcal M^*(s_0)=\mathcal T^*$. We generate $20\times30\times40\times50$ tensors having simultaneous LR, SP, and AR structure, with all four substantive modes retained in the recovered tensor. The Tucker construction is detailed in Supplementary Material. We compare HM-ADMM with Gandy-full at $\rho\in\{0.20,0.40,0.60,0.80\}$, again using 100 replications at each rate. The regularization parameters of HM-ADMM are selected by cross-validation, with the AR penalty using a fixed estimate of $\phi$ obtained from adjacent observed pairs in the fitting set.

Table~\ref{tab:sim_nosmooth_mse} shows that HM-ADMM has lower MSE throughout, with its advantage increasing as the observation rate rises.  Thus, joint heterogeneous penalties remain effective in the singleton-index degeneration, where no local borrowing occurs.

\begin{table}[!h]
\centering
\caption{Mean MSE in the singleton-index LR+SP+AR design without local borrowing.}
\label{tab:sim_nosmooth_mse}
\fontsize{8}{9.6}\selectfont
\begin{tabular}{cccc}
\toprule
$\rho$ & HM-ADMM & Gandy-full & Relative reduction \\
\midrule
0.20 & 20.024 & 20.132 & 0.5\% \\
0.40 & 13.127 & 15.289 & 14.2\% \\
0.60 &  7.360 & 10.457 & 29.6\% \\
0.80 &  2.797 &  5.614 & 50.2\% \\
\bottomrule
\end{tabular}
\end{table}

\section{Real-Data Applications}
\label{sec:realdata}

\subsection{CMEMS biogeochemical data}
\label{subsec:bgc}

We use 24 monthly fields from 2024--2025 over the Northwest Pacific domain \(120^\circ\mathrm{E}\text{--}150^\circ\mathrm{E}\), \(10^\circ\mathrm{N}\text{--}40^\circ\mathrm{N}\), from the Copernicus Marine product~\cite{eu2021global}\footnote{Copernicus Marine Service, \emph{GLOBAL ANALYSISFORECAST BGC 001 028}, \url{https://data.marine.copernicus.eu/product/GLOBAL_ANALYSISFORECAST_BGC_001_028/description}.}. For each of 100 seeds, a contiguous \(50\times50\) spatial crop forms a \(50\times50\times24\times13\times50\) tensor (longitude$\times$latitude$\times$time$\times$variable$\times$depth). The 13 variables cover carbon, \(\mathrm{CO}_2\), nutrients, and optical properties. Land masks and reduced coverage at greater depths induce structured missingness; invalid original entries are excluded from scoring. We split the remaining entries equally between training and testing and use identical crops and masks across methods, yielding \(1{,}588{,}656{,}264\) paired test entries. Longitude and latitude are treated as smooth indexing modes, while time, variable, and depth are modeled by low-rank structure, with an alternative specification that assigns AR structure to time.

Each variable--depth channel is standardized using its training mean and standard deviation; values below \(1\%\) of the median valid standard deviation for that variable are replaced by this floor. This prevents nearly constant deep channels from being excessively amplified and makes pooled errors more comparable across physical units. We report MAE, correlation, and \( \operatorname{NRMSE} =\left\{ {\sum_{\Omega_{\mathrm{test}}}(\widehat Z-Z)^2}/ {\sum_{\Omega_{\mathrm{test}}}Z^2} \right\}^{1/2} \) on both the standardized and original-unit scales.

Both HM specifications use Epanechnikov weighting over the two spatial coordinates. HM-LR3 assigns LR structure to time, variable, and depth, whereas HM-LR2\_AR replaces temporal LR structure with AR regularization, with $\phi$ pre-estimated from adjacent pairs in the fitting observations. We compare them with slicewise and full-tensor versions of Gandy and Takeuchi-ATF. Hyperparameters minimize pooled standardized NRMSE on five separate validation crops, using an \(80\%/20\%\) fit--validation split within each outer training set.

Table~\ref{tab:bgc_main} shows that, on the standardized scale, HM-LR2\_AR has the lowest NRMSE and MAE, while HM-LR3 has the highest correlation. HM-LR2\_AR reduces NRMSE by \(42.8\%\) relative to the strongest non-HM competitor, Gandy-slicewise. The HM variants also lead on the original-unit scale, although these pooled metrics are dominated by variables with larger physical scales.

\begin{table}[!h]
\centering
\caption{CMEMS BGC results pooled over the held-out entries from 100 crops, reported on both the standardized and original-unit scales.}
\label{tab:bgc_main}
\fontsize{8}{9.6}\selectfont
\begin{tabular}{lrrrrrr}
\toprule
& \multicolumn{3}{c}{Standardized scale} & \multicolumn{3}{c}{Original-unit scale} \\
\cmidrule(lr){2-4} \cmidrule(lr){5-7}
Method & NRMSE & MAE & Corr & NRMSE & MAE & Corr \\
\midrule
HM-LR3                  & 0.1898 & 0.0631          & \textbf{0.9822} & \textbf{0.01116} & 0.08311          & \textbf{0.9999293} \\
HM-LR2\_AR & \textbf{0.1897} & \textbf{0.0620} & 0.9821 & 0.01117 & \textbf{0.08160} & 0.9999291 \\
Gandy (slicewise)       & 0.3318          & 0.2149          & 0.9443          & 0.01699          & 0.26243          & 0.9998361 \\
Gandy (full)            & 0.4259          & 0.2609          & 0.9068          & 0.02324          & 0.32447          & 0.9996932 \\
Takeuchi-ATF (slicewise) & 0.4209       & 0.2644          & 0.9075          & 0.02210          & 0.32398          & 0.9997227 \\
Takeuchi-ATF (full)   & 0.6973          & 0.4459          & 0.7168          & 0.04011          & 0.56972          & 0.9990862 \\
\bottomrule
\end{tabular}
\end{table}

\subsection{Fresh Lung FS-FLIM data}
\label{subsec:fresh_lung}

The Fresh Lung fluorescence spectral-lifetime imaging (FS-FLIM) dataset
contains 20 public tiles~\citep{williams2021fsflimdataset}\footnote{\href{https://datashare.ed.ac.uk/items/ca312ea4-7277-4339-a7e4-25432268a0be}{DataShare record}; \url{https://doi.org/10.7488/ds/3099}.} of size \(256\times256\times16\times512\) (row, column, lifetime, wavelength). At each wavelength, we subtract the spatial mean of the last lifetime bin from the first 15 bins and set negative values to zero. In each ROI, we sum the first 384 wavelength channels in groups of three and swap the last two modes, yielding a \(64\times64\times128\times15\) tensor (row, column, wavelength, lifetime). Each ROI is divided by its \(99.9\)th percentile and clipped to \([0,1]\). We use spatial smoothing, graph regularization for wavelength, and low-rank regularization for lifetime.

Fixed ROIs from distinct tiles are reserved for tuning. The final comparison uses 100 independently drawn \(64\times64\) ROIs at \(\mathrm{SR}\in\{2\%,5\%,10\%,20\%\}\). All methods use the same observation mask for each ROI and sampling rate, and MSE, MPSNR, MSSIM, and ERGAS are evaluated on unobserved entries after predictions are clipped to \([0,1]\).

HM-ADMM uses a radial symmetric Epanechnikov kernel over the two spatial coordinates, a GEO penalty over wavelength, and an LR penalty over lifetime. Tian-MNTC and Gandy-full use the same training and test entries; Gandy-full is applied to the complete four-dimensional tensor and capped at 500 iterations. The HM neighborhood and regularization parameters are selected before the 100-ROI evaluation.

\begin{figure}[!h]
\centering
\includegraphics[width=0.8\textwidth,keepaspectratio]{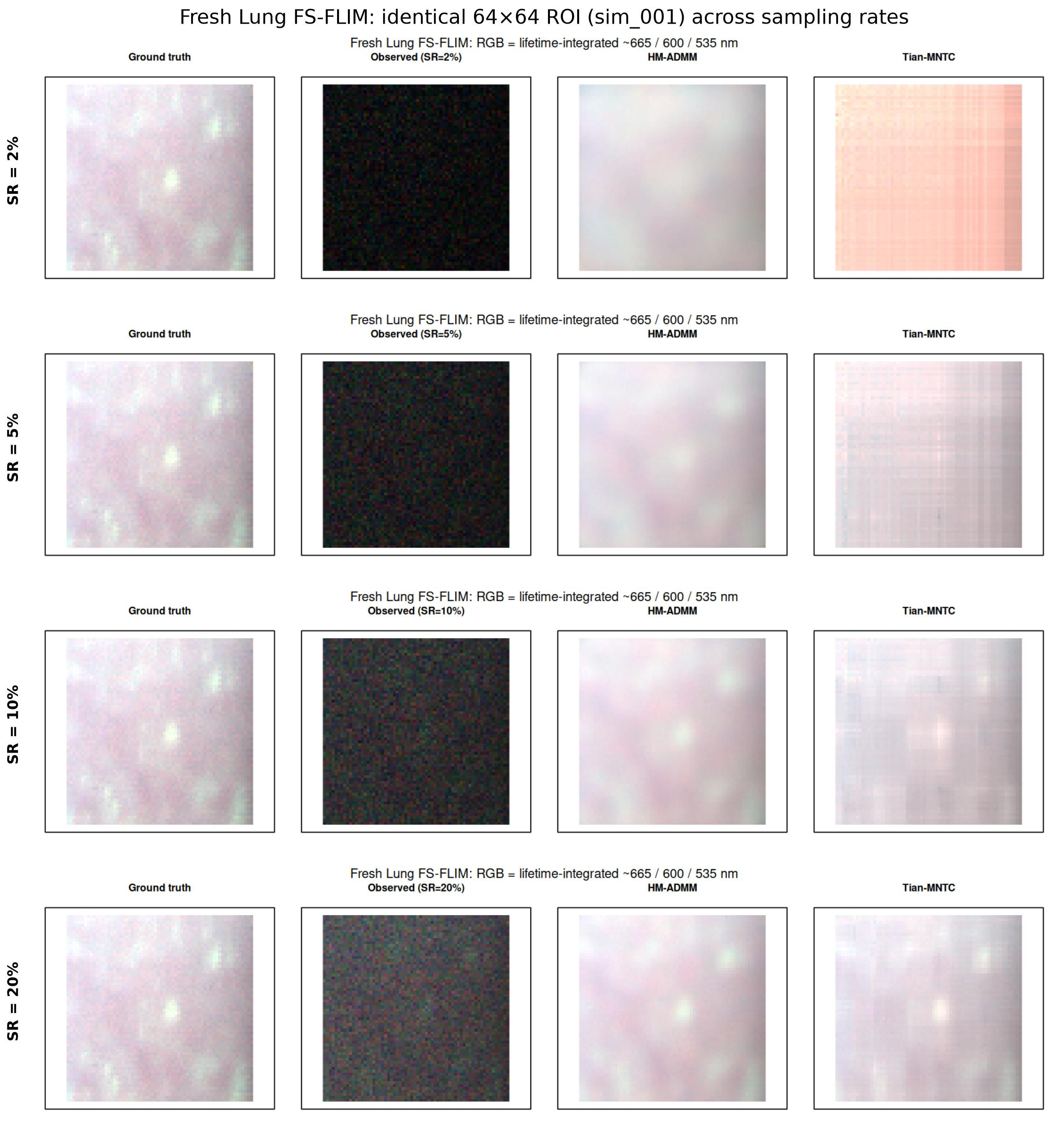}
\caption{Fresh Lung FS-FLIM reconstruction for the same $64\times64$ ROI across sampling rates.  The RGB display integrates lifetime intensity at approximately 665, 600, and 535\,nm.}
\label{fig:fresh_lung_sr}
\end{figure}

Table~\ref{tab:fresh_lung_sr} and Figure~\ref{fig:fresh_lung_sr} show the clearest separation under extreme undersampling. Relative to Tian-MNTC, HM-ADMM reduces missing-entry MSE by \(69.6\%\) at \(2\%\) sampling and by \(25.4\%\) at \(5\%\), while running approximately \(9.3\) and \(11.6\) times faster, respectively. The corresponding gains in MPSNR, MSSIM, and ERGAS are consistent with improved structural recovery.

\begin{table}[!h]
\centering
\caption{Fresh Lung FS-FLIM completion over 100 independently drawn ROIs. Entries are mean $\pm$ standard deviation. Gandy-full uses the complete four-dimensional tensor and is capped at 500 iterations.}
\label{tab:fresh_lung_sr}
\resizebox{\textwidth}{!}{%
\begin{tabular}{clccccc}
\toprule
SR & Method & MSE & MPSNR~$\uparrow$ & MSSIM~$\uparrow$ & ERGAS~$\downarrow$ & Runtime (s) \\
\midrule
$2\%$  & HM-ADMM   & $\mathbf{0.000698\pm0.000241}$ & $\mathbf{35.075\pm1.782}$ & $\mathbf{0.9070\pm0.0602}$ & $\mathbf{51.527\pm10.349}$ & $\mathbf{133.1\pm26.6}$ \\
       & Tian-MNTC & $0.002296\pm0.000438$ & $30.131\pm0.856$ & $0.7410\pm0.0690$ & $89.425\pm9.898$ & $1240.1\pm342.5$ \\
       & Gandy-full & $0.019127\pm0.006246$ & $22.509\pm1.529$ & $0.4234\pm0.0290$ & $175.246\pm14.230$ & $8318\pm921.1$ \\
\addlinespace
$5\%$  & HM-ADMM   & $\mathbf{0.000499\pm0.000224}$ & $\mathbf{36.354\pm2.116}$ & $\mathbf{0.9233\pm0.0590}$ & $\mathbf{44.212\pm9.341}$ & $\mathbf{116.2\pm21.8}$ \\
       & Tian-MNTC & $0.000669\pm0.000260$ & $35.307\pm1.692$ & $0.9043\pm0.0614$ & $47.411\pm7.634$ & $1349.9\pm360.2$ \\
       & Gandy-full & $0.014290\pm0.005133$ & $24.571\pm1.755$ & $0.5960\pm0.0388$ & $138.097\pm11.748$ & $8438\pm1070.3$ \\
\addlinespace
$10\%$ & HM-ADMM   & $0.000413\pm0.000216$ & $37.034\pm2.323$ & $\mathbf{0.9331\pm0.0551}$ & $41.710\pm9.082$ & $\mathbf{111.8\pm10.0}$ \\
       & Tian-MNTC & $\mathbf{0.000376\pm0.000230}$ & $\mathbf{37.509\pm2.466}$ & $0.9300\pm0.0569$ & $\mathbf{39.975\pm8.622}$ & $875.2\pm110.8$ \\
       & Gandy-full & $0.008946\pm0.003793$ & $27.289\pm2.052$ & $0.7459\pm0.0558$ & $101.950\pm8.760$ & $8226\pm1030.3$ \\
\addlinespace
$20\%$ & HM-ADMM   & $0.000345\pm0.000212$ & $37.701\pm2.574$ & $0.9448\pm0.0479$ & $39.843\pm9.057$ & $\mathbf{94.3\pm16.0}$ \\
       & Tian-MNTC & $\mathbf{0.000299\pm0.000215}$ & $\mathbf{38.416\pm2.855}$ & $\mathbf{0.9458\pm0.0481}$ & $\mathbf{37.875\pm9.058}$ & $491.2\pm168.2$ \\
       & Gandy-full & $0.003282\pm0.001960$ & $31.694\pm2.641$ & $0.8812\pm0.0637$ & $62.392\pm8.714$ & $7982\pm1096.8$ \\
\bottomrule
\end{tabular}%
}
\end{table}

The results support the intended mechanism: local aggregation supplies information from nearby pixels when few entries are observed at the target, while the graph penalty stabilizes neighboring wavelength channels. At \(10\%\) and \(20\%\) sampling, Tian-MNTC has lower pointwise error, whereas HM-ADMM remains faster and has comparable structural similarity. Gandy-full is less accurate throughout, with MSE reductions of \(89.5\%\)--\(96.5\%\) for HM-ADMM and runtimes \(63\)--\(85\) times larger; all Gandy runs reach the 500-iteration cap.

\section{Discussion}
\label{sec:discussion}

This work develops a unified framework for tensor recovery when different modes exhibit different forms of regularity. Local weighting borrows information across a smooth indexing domain, while mode-specific penalties encode low-rank, group-sparse, geometry-aware, and autoregressive structure within the retained tensor. The resulting consensus HM-ADMM algorithm separates these components into tractable updates. For the population-quadratic estimator, the recovery theory distinguishes stochastic fluctuation, rank and support complexity, smoothing bias, and quadratic structural mismatch, and yields optimal bandwidth orders and normalized rates for uniform tensor completion. It also identifies how restricted quadratic curvature and structural mismatch determine whether GEO/AR regularization strictly sharpens the rank- and support-dependent Frobenius guarantee.

The current theory assumes independent trace-regression observations, a common population design operator within each local window, and an interior target under a regular fixed-domain design. Future work may allow the design operator to vary smoothly across locations, accommodate dependent or nonuniform sampling, and develop boundary-adaptive smoothing. Other directions include jointly learning graph operators and autoregressive parameters and providing theoretical guarantees for data-driven bandwidth and structural selection. These extensions would broaden the framework while retaining its separation between local information borrowing and mode-specific regularization.

\phantomsection\label{supplementary-material}
\bigskip
\begin{center}
{\large\bf SUPPLEMENTARY MATERIAL}
\end{center}
\begin{description}
\item[Technical supplement:]
Additional deterministic specializations and proofs for the independent trace-regression results. (PDF)
\end{description}

\bibliographystyle{agsm}
\bibliography{bibliography}
\end{document}